\documentclass[10pt,twocolumn,letterpaper]{article}

\usepackage{wacv}              

\usepackage{graphicx}
\usepackage{amsmath}
\usepackage{amssymb}
\usepackage{booktabs}

\usepackage{booktabs}
\usepackage{tabularx}
\usepackage{subcaption}
\usepackage{comment}
\usepackage{enumitem}
\usepackage{multirow}

\usepackage{tikz}
\usetikzlibrary{positioning}
\usepackage{xcolor}
\usepackage{algorithm2e}
\usetikzlibrary{decorations.pathreplacing,calligraphy, arrows}

\definecolor{stdSampler}{RGB}{215, 25, 28}
\definecolor{mscSampler}{RGB}{68,105,188}
\definecolor{mscvbswcSampler}{RGB}{105,226,68}

\usepackage{amsmath,amsfonts,bm}

\def\eqref#1{equation~\ref{#1}}

\def\1{\bm{1}}

\DeclareMathAlphabet{\mathsfit}{\encodingdefault}{\sfdefault}{m}{sl}
\SetMathAlphabet{\mathsfit}{bold}{\encodingdefault}{\sfdefault}{bx}{n}

\def\gB{{\mathcal{B}}}

\def\gD{{\mathcal{D}}}
\def\gE{{\mathcal{E}}}

\def\gH{{\mathcal{H}}}

\def\gN{{\mathcal{N}}}

\def\gS{{\mathcal{S}}}

\newcommand{\gBs}{\left|\gB\right|}

\usepackage[pagebackref,breaklinks,colorlinks]{hyperref}

\usepackage[capitalize]{cleveref}
\crefname{section}{Sec.}{Secs.}
\Crefname{section}{Section}{Sections}
\Crefname{table}{Table}{Tables}
\crefname{table}{Tab.}{Tabs.}

\def\confName{WACV}
\def\confYear{2024}

\begin{document}

\title{On the Efficacy of Multi-scale Data Samplers for Vision Applications}

\author{Elvis Nunez\thanks{Work done during an internship at Apple.}\\
University of California, Los Angeles\\
{\tt\small elvis.nunez@ucla.edu}
\and
Thomas Merth\\
Apple\\
{\tt\small tmerth@apple.com}
\and
Anish Prabhu\thanks{Work done at Apple.} \\
Apple\\
{\tt\small prabhuanish97@gmail.com}
\and
Mehrdad Farajtabar \\
Apple\\
{\tt\small farajtabar@apple.com}
\and
Mohammad Rastegari \\
Apple\\
{\tt\small mrastegari@apple.com}
\and
Sachin Mehta \\
Apple\\
{\tt\small sachin\_mehta@apple.com}
\and
Maxwell Horton \\
Apple\\
{\tt\small mchorton@apple.com}
}

\maketitle

\begin{abstract}
Multi-scale resolution training has seen an increased adoption across multiple vision tasks, including classification and detection. Training with smaller resolutions enables faster training at the expense of a drop in accuracy. Conversely, training with larger resolutions has been shown to improve performance, but memory constraints often make this infeasible. In this paper, we empirically study the properties of multi-scale training procedures. We focus on variable batch size multi-scale data samplers that randomly sample an input resolution at each training iteration and dynamically adjust their batch size according to the resolution. Such samplers have been shown to improve model accuracy beyond standard training with a fixed batch size and resolution, though it is not clear why this is the case. We explore the properties of these data samplers by performing extensive experiments on ResNet-101 and validate our conclusions across multiple architectures, tasks, and datasets. We show that multi-scale samplers behave as implicit data regularizers and accelerate training speed. Compared to models trained with single-scale samplers, we show that models trained with multi-scale samplers retain or improve accuracy, while being better-calibrated and more robust to scaling and data distribution shifts. We additionally extend a multi-scale variable batch sampler with a simple curriculum that progressively grows resolutions throughout training, allowing for a compute reduction of more than 30\%. We show that the benefits of multi-scale training extend to detection and instance segmentation tasks, where we observe a 37\% reduction in training FLOPs along with a 3-4\% mAP increase on MS-COCO using a Mask R-CNN model. 
\end{abstract}

\section{Introduction}
\label{sec:intro}

Training neural networks for visual recognition has traditionally focused on using a fixed input resolution throughout training (known as \textit{single-scale training}) \cite{alexnet,resnet,VGG}. However, training at multiple image resolutions (known as \textit{multi-scale training}) may offer some advantages. For example, expanding training or fine-tuning to include variable input sizes has been shown to improve accuracy in object detection~\cite{yolo9000}. Moreover, multi-scale training has the potential to result in faster training by utilizing available hardware more efficiently~\cite{cvnets,MobileViT}. Despite the aforementioned works, multi-scale samplers remain largely unexplored. Most previous works that use multi-scale resolution training~\cite{yolo9000,cvnets,MobileViT} present only accuracy and run-time numbers, and lack a systematic analysis of how these samplers compare to conventional samplers on other important metrics. In particular, the effect that multi-scale training has on the generalization, robustness, and training convergence of the final models has not been extensively studied.

In this work, we provide a comprehensive empirical study of models trained with multi-scale samplers. We show that by dynamically adjusting the batch size and input resolution on different GPUs, one can achieve accelerated training of deep neural networks while preserving or improving the model accuracy. In our analysis, we study four different sampling strategies. The first corresponds to standard single-scale training with a fixed batch size (SSc-FBS). The second is a multi-scale sampler with a fixed batch size (MSc-FBS) \cite{yolo9000}. The third sampler corresponds to multi-scale training with a variable batch size (MSc-VBS)~\cite{cvnets}, in which batch sizes are dynamically adjusted so that batch sizes are large when training with a small resolution, and small when training with a large resolution. Following curriculum and progressive learning \cite{curriculum_learning,tan2021efficientnetv2}, we also extend the MSc-VBS sampler by adding a curriculum that slowly increases the resolutions of training images throughout the course of training. We call this the multi-scale variable batch size with curriculum (\textit{MSc-VBSWC}) sampler. MSc-VBS and MSc-VBSWC reduce the training FLOPs of ResNet-101 by 23\% and 30\%, respectively.

We study the impact of the choice of sampling strategy on the training speed, robustness, and regularization. In studying training speed, we find that multi-scale training can be used to reduce training FLOPs for ResNet-101 \cite{resnet} by up to $30\%$ without sacrificing accuracy. In studying robustness, we find that training with multi-scale resolution improves robustness to changes in input resolution as well as performance on often-misclassified examples from ImageNet-A (\cite{ImageNet-A}).  We also demonstrate that multi-scale samplers result in models with lower calibration errors, signifying that these models' predictions are better correlated with the probability that the predictions are correct.  We further show that embeddings learned by the multi-scale samplers exhibit lower variance which leads to faster training.  In studying regularization, we find that multi-scale samplers reduce the dependence on other hand-tuned regularizers such as stochastic dropout \cite{stochastic_depth}, suggesting that multi-scale samplers act as regularizers. Finally, we study the benefits of multi-scale samplers for different architectures and tasks. We report notable computation reductions in training with EfficientNet-B3 \cite{efficientnet}, and RegNetY-16GF \cite{regnet}. We also show that multi-scale samplers results in faster training and significant accuracy improvements when performing instance segmentation and detection with Mask R-CNN~\cite{mask_rcnn} on the MS-COCO dataset~\cite{ms_coco}.

In summary, our contributions are: 
(1) We present an empirical analysis of different data sampling strategies for vision tasks. 
(2) We propose an improvement on standard multi-scale samplers by enforcing a curriculum on the batch size and resolutions.
(3) We demonstrate improved accuracy and reduced training time of the multi-scale samplers and analyze their properties from the perspective of training speed, robustness, and regularization. 
(4) We extend our analysis to other network architectures, vision tasks, and datasets.

\section{Related work}

\paragraph{Multi-Scale Training}
Training and inference with heterogeneous input resolutions has been used for a variety of purposes in computer vision. In object detection, image classifier backbones are often fine-tuned at resolutions higher than they were trained~\cite{yolo9000}; a similar procedure is found in vision transformer training \cite{DeiT}, where a pre-trained classifier is fine-tuned at multiple resolutions. The authors of \cite{FixRes} demonstrate that fine-tuning at the final evaluation resolution can reduce the train-test domain gap and increase accuracy.

The idea of training architectures on smaller inputs to increase training and inference speed has been studied for the purpose of efficient architecture design. MobileNets \cite{Howard2017MobileNetsEC,Sandler2018MobileNetV2IR,Howard2019SearchingFM} include a variety of image input sizes, allowing a variety of efficiency-accuracy trade-offs. The authors of \cite{progressive-resizing-howard2018} consider progressive resizing, where input resolutions are gradually increased throughout training. Building upon \cite{progressive-resizing-howard2018}, EfficientNetv2 \cite{tan2021efficientnetv2} augments progressive resizing by modulating the strength of the regularization employed at different training resolutions. These works suggest that progressive resizing can be beneficial, although little analysis is shown to provide justification for their effectiveness.

\paragraph{Regularization}
Model regularization by manipulating model inputs has been extensively studied through the lens of data augmentations, which typically include cropping, rotation, color modulation, reflection, and more \cite{Cubuk2018AutoAugmentLA,Cubuk2019RandaugmentPA,Mller2021TrivialAugmentTY,LingChen2020UniformAugmentAS,Lemley2017SmartAL,Zhang2019AdversarialA,GoogLeNet, ghiasi2021simple}. Data augmentation has been shown to improve generalization \cite{Yang2022ImageDA} as well as final validation accuracy. However, existing works that have studied multi-scale training generally do not thoroughly investigate its regularization properties with the same thoroughness as augmentation strategies.

\paragraph{Robustness}
Previous works (e.g., \cite{MixMatch, MobileViT, yolo9000, deeplabv3, pspnet}) propose training with different  resolutions at train-time. Generally, these works demonstrate that the resulting models are more robust to a wider range of resolutions at test-time. Robustness to other domain shifts (such as ImageNet-A \cite{ImageNet-A}, ImageNet-C \cite{ImageNet-C}, ImageNet-R \cite{ImageNet-R}), and analyses of calibration properties \cite{ece_calibration} have not been studied.

Whereas previous works narrowly explore the accuracy and speed/memory benefits of multi-scale training, we aim to explore the benefits of multi-scale training, including robustness and calibration, across a multitude of architectures, tasks, and datasets. 
\section{Overview of Image Data Samplers}
\label{sec:overview}
When training a deep neural network $\gN$ for vision tasks, gradient descent is performed over a dataset $\gD$ for a specific number of epochs, $E$. Throughout training, $\gN$ is fed batches $\gB \subset \gD$ with shape $\gBs \triangleq (B, C, H, W)$ where $B$ denotes the batch size, $C$ is the number of input channels, and $H,W$ are the image height and width, respectively. Typically, the batch shape $\gBs$ is fixed throughout training so that at each iteration, the model is fed a fixed number of samples at a fixed resolution. Throughout this paper, we refer to the protocol that fetches data batches at each training iteration as the \textit{data sampler}. We discuss several data sampling strategies and their trade-offs below. 

\begin{figure}[t!]
    \begin{minipage}{.5\linewidth}
    \centering
    \resizebox{!}{100px}{\hspace{-125pt}\definecolor{patch0}{RGB}{127,201,127}
\definecolor{patch1}{RGB}{190,174,212}
\definecolor{patch2}{RGB}{253,192,134}
\definecolor{patch3}{RGB}{255,255,153}

\definecolor{patch4}{RGB}{56,108,176}
\definecolor{patch5}{RGB}{240,2,127}
\definecolor{patch6}{RGB}{191,91,23}
\definecolor{patch7}{RGB}{158,1,66}

\definecolor{patch8}{RGB}{228,26,28}
\definecolor{patch9}{RGB}{55,126,184}
\definecolor{patch10}{RGB}{77,175,74}
\definecolor{patch11}{RGB}{152,78,163}

\definecolor{patch12}{RGB}{255,127,0}
\definecolor{patch13}{RGB}{255,255,51}
\definecolor{patch14}{RGB}{166,86,40}
\definecolor{patch15}{RGB}{247,129,191}

\newcommand{\drawCube}[5]{
    \begin{tikzpicture}[line join=round]
      \pgfmathsetmacro{\cubex}{#1}
      \pgfmathsetmacro{\cubey}{#2}
      \pgfmathsetmacro{\cubez}{#3}
      \draw[draw=black, fill=#4, opacity=1] (0,0,0)-- ++(-\cubex,0,0) -- ++(0,-\cubey,0) -- ++(\cubex,0,0) -- cycle
       (0,0,0) -- ++(0,0,-\cubez) -- ++(0,-\cubey,0) -- ++(0,0,\cubez) -- cycle
       (0,0,0) -- ++(-\cubex,0,0) -- ++(0,0,-\cubez) -- ++(\cubex,0,0) -- cycle;
    \end{tikzpicture}
}

\newcommand{\drawCubeDiv}[5]{
    \begin{tikzpicture}[line join=round]
      \pgfmathsetmacro{\cubex}{#1}
      \pgfmathsetmacro{\cubey}{#2}
      \pgfmathsetmacro{\cubez}{#3}
      \draw[draw=black, opacity=1] (0,0,0)-- ++(-\cubex,0,0) -- ++(0,-\cubey,0) -- ++(\cubex,0,0) -- cycle
       (0,0,0) -- ++(0,0,-\cubez) -- ++(0,-\cubey,0) -- ++(0,0,\cubez) -- cycle
       (0,0,0) -- ++(-\cubex,0,0) -- ++(0,0,-\cubez) -- ++(\cubex,0,0) -- cycle;
       
       \draw[draw=black, opacity=1] (0,0,0)-- ++(-\cubex,0,0) -- ++(0,-\cubey,0) -- ++(\cubex,0,0) -- cycle
       (0,0,0) -- ++(0,0,-1) -- ++(0,-\cubey,0) -- ++(0,0,1) -- cycle
       (0,0,0) -- ++(-\cubex,0,0) -- ++(0,0,-1) -- ++(\cubex,0,0) -- cycle;
    \end{tikzpicture}
}

\newcommand{\drawCubeEmpty}[5]{
    \begin{tikzpicture}[line join=round]
      \pgfmathsetmacro{\cubex}{#1}
      \pgfmathsetmacro{\cubey}{#2}
      \pgfmathsetmacro{\cubez}{#3}
      \draw[draw=white, opacity=1] (0,0,0)-- ++(-\cubex,0,0) -- ++(0,-\cubey,0) -- ++(\cubex,0,0) -- cycle
       (0,0,0) -- ++(0,0,-\cubez) -- ++(0,-\cubey,0) -- ++(0,0,\cubez) -- cycle
       (0,0,0) -- ++(-\cubex,0,0) -- ++(0,0,-\cubez) -- ++(\cubex,0,0) -- cycle;
    \end{tikzpicture}
}

\tikzset{
    cross/.pic = {
    \draw[rotate = 45] (-#1,0) -- (#1,0);
    \draw[rotate = 45] (0,-#1) -- (0, #1);
    }
}

\newcommand{\stdDDP}{
    \begin{tikzpicture}
        \node[label={[shift={(-2.8,-2.2)}]\scalebox{2.5}{GPU-4}}] (tensor_a) at (0, 0, 0) {\drawCube{2}{2}{2}{red!30}{1.0}};
        \node (tensor_b) at (1, 0, 0) {\drawCube{0.1}{2}{2}{yellow!20}{1.0}};
        \node (tensor_b) at (2.05, 0, 0) {\drawCube{2}{2}{2}{red!30}{1.0}};
        \node (tensor_b) at (3.1, 0, 0) {\drawCube{0.1}{2}{2}{yellow!20}{1.0}};
        
        \node[label={[shift={(-2.8,-2.2)}]\scalebox{2.5}{GPU-3}}] (tensor_a1) at (0, 2, 0) {\drawCube{2}{2}{2}{red!30}{1.0}};
        \node (tensor_b) at (1, 2, 0) {\drawCube{0.1}{2}{2}{yellow!20}{1.0}};
        \node (tensor_b) at (2.05, 2, 0) {\drawCube{2}{2}{2}{red!30}{1.0}};
        \node (tensor_b) at (3.1, 2, 0) {\drawCube{0.1}{2}{2}{yellow!20}{1.0}};
        
        \node[label={[shift={(-2.8,-2.2)}]\scalebox{2.5}{GPU-2}}] (tensor_a2) at (0, 4, 0) {\drawCube{2}{2}{2}{red!30}{1.0}};
        \node (tensor_b) at (1, 4, 0) {\drawCube{0.1}{2}{2}{yellow!20}{1.0}};
        \node (tensor_b) at (2.05, 4, 0) {\drawCube{2}{2}{2}{red!30}{1.0}};
        \node (tensor_b) at (3.1, 4, 0) {\drawCube{0.1}{2}{2}{yellow!20}{1.0}};
        
        \node[label={[shift={(-2.8,-2.2)}]\scalebox{2.5}{GPU-1}}] (tensor_a3) at (0, 6, 0) {\drawCube{2}{2}{2}{red!30}{1.0}};
        \node (tensor_b) at (1, 6, 0) {\drawCube{0.1}{2}{2}{yellow!20}{1.0}};
        \node (tensor_b) at (2.05, 6, 0) {\drawCube{2}{2}{2}{red!30}{1.0}};
        \node (tensor_c) at (3.1, 6, 0) {\drawCube{0.1}{2}{2}{yellow!20}{1.0}};
        
        \draw[->, line width=0.5mm] (-1.4, -1.4) -- (4, -1.4);
        \node at (3.5, -1.9)  {\scalebox{2.5}{time}};
        
        \node[above left=0.75cm and -5cm of tensor_a3, align=center] (b){\scalebox{3}{\textcolor{stdSampler}{\bfseries SSc-FBS}}};
    
    \end{tikzpicture}
}

\newcommand{\mscvbswcDDP}{
    \begin{tikzpicture}
        \node[label={[shift={(-6.25,-1.8)}]\scalebox{2.5}{GPU-4}}] (tensor_a) at (0, 0, 0) {};
        \node at (0, 0, 3) {\drawCube{7}{0.5}{0.5}{black!10!green}{1.0}}; 
        \node[label={[shift={(-6.25,0.4)}]\scalebox{2.5}{GPU-3}}] (tensor_b) at (0, 0, 0) {};
        \node at (-1.15, 2, 2.5) {\drawCube{5}{0.75}{0.75}{black!10!green}{1.0}}; 
        \node[label={[shift={(-6.25,2.6)}]\scalebox{2.5}{GPU-2}}] (tensor_c) at (0, 0, 0) {};
        \node at (-2.4, 4.01, 1.5) {\drawCube{3}{1.25}{1.25}{black!10!green}{1.0}}; 
        \node[label={[shift={(-6.25,3.8)}]\scalebox{2.5}{GPU-1}}] (tensor_d) at (0, 0, 0) {};
        \node at (-1, 6, 4) {\drawCube{4}{1}{1}{black!10!green}{1.0}}; 
        
        \node at (3.2-1.5+1.1, -0.2, 2) {\drawCube{0.1}{2}{2}{yellow!20}{1.0}};
        \node at (3.2-1.5+1.1, 1.8, 2) {\drawCube{0.1}{2}{2}{yellow!20}{1.0}};
        \node at (3.2-1.5+1.1, 3.8, 2) {\drawCube{0.1}{2}{2}{yellow!20}{1.0}};
        \node (marked) at (3.2-1.5+1.1, 5.8, 2) {\drawCube{0.1}{2}{2}{yellow!20}{1.0}};
        
        \node at (5.40-1.95+3.4-2.3, 0.5, 2.5) {\drawCube{3}{1}{2}{black!10!green}{1.0}};
        \node at (4.44-1.95+3.4-2.3, 2.02, 1.6) {\drawCube{1}{2}{4}{black!10!green}{1.0}};
        \node at (5.42-1.95+3.4-2.3, 4.01, 3.05) {\drawCube{3}{2}{1}{black!10!green}{1.0}};
        \node at (5.45-1.95+3.4-2.3, 6.55, 3.95) {\drawCube{2}{2}{2}{black!10!green}{1.0}};
        
        \node at (8.2-2.3, -0.2, 2) {\drawCube{0.1}{2}{2}{yellow!20}{1.0}};
        \node at (8.2-2.3, 1.8, 2) {\drawCube{0.1}{2}{2}{yellow!20}{1.0}};
        \node at (8.2-2.3, 3.8, 2) {\drawCube{0.1}{2}{2}{yellow!20}{1.0}};
        \node at (8.2-2.3, 5.8, 2) {\drawCube{0.1}{2}{2}{yellow!20}{1.0}};

        \draw[->, line width=0.75mm] (-4.75, -2.35) -- (7, -2.35);
        
        \node at (6.5, -2.75)  {\scalebox{2.5}{time}};
        \node[above left=0.75cm and -3cm of marked, align=center] (b) {\scalebox{3}{\textcolor{mscvbswcSampler}{\bfseries MSc-VBSWC}}};
    
    \end{tikzpicture}
}

\newcommand{\mscDDP}{
    \begin{tikzpicture}
    
        \node[label={[shift={(-3.6,-1.75)}]\scalebox{2.5}{GPU-4}}] (tensor_a) at (0.92, 0, 3) {\drawCube{4}{2}{1}{blue!30}{1.0}};
        \node[label={[shift={(-2.85,-2.25)}]\scalebox{2.5}{GPU-3}}] at (-0.08, 2, 2.5) {\drawCube{2}{2}{2}{blue!30}{1.0}};
        \node[label={[shift={(-2.85,-3.0)}]\scalebox{2.5}{GPU-2}}] at (-0.57, 4.01, 1.5) {\drawCube{1}{2}{4}{blue!30}{1.0}};
        \node[label={[shift={(-3.85,-1.0)}]\scalebox{2.5}{GPU-1}}] at (1.5, 6.1, 4) {\drawCube{4}{1}{2}{blue!30}{1.0}};
        
        \node at (3.2-0.5, -0.2, 2) {\drawCube{0.1}{2}{2}{yellow!20}{1.0}};
        \node at (3.2-0.5, 1.8, 2) {\drawCube{0.1}{2}{2}{yellow!20}{1.0}};
        \node at (3.2-0.5, 3.8, 2) {\drawCube{0.1}{2}{2}{yellow!20}{1.0}};
        \node (marked) at (3.2-0.5, 5.8, 2) {\drawCube{0.1}{2}{2}{yellow!20}{1.0}};
        
        \node at (5.90-0.95, 0.5, 2.5) {\drawCube{4}{1}{2}{blue!30}{1.0}};
        \node at (4.44-0.95, 2.02, 1.6) {\drawCube{1}{2}{4}{blue!30}{1.0}};
        \node at (5.92-0.95, 4.01, 3.05) {\drawCube{4}{2}{1}{blue!30}{1.0}};
        \node at (5.45-0.95, 6.55, 3.95) {\drawCube{2}{2}{2}{blue!30}{1.0}};
        
        \node at (8.2-1.4, -0.2, 2) {\drawCube{0.1}{2}{2}{yellow!20}{1.0}};
        \node at (8.2-1.4, 1.8, 2) {\drawCube{0.1}{2}{2}{yellow!20}{1.0}};
        \node at (8.2-1.4, 3.8, 2) {\drawCube{0.1}{2}{2}{yellow!20}{1.0}};
        \node at (8.2-1.4, 5.8, 2) {\drawCube{0.1}{2}{2}{yellow!20}{1.0}};
        
        \draw[->, line width=0.5mm] (-2.45, -2.35) -- (7, -2.35);
        
        \node at (6.5, -2.75)  {\scalebox{2.5}{time}};
        \node[above=0.75cm of marked, align=center] (b) {\scalebox{3}{\textcolor{mscSampler}{\bfseries MSc-VBS}}};
    
    \end{tikzpicture}
}

\newcommand{\tensorNotations}{
    \begin{tikzpicture}
        \node[label={[shift={(-1.75,-2.5)}]\rotatebox{90}{ \scalebox{1.35}{Height}}}, label={[shift={(-1.2,-0.9)}]\rotatebox{50}{\scalebox{1.35}{Width}}}, label={[shift={(-0.5,-3.5)}] \scalebox{1.35}{Batch}}] (tensor_c) {\drawCube{2}{2}{2}{gray!40}{1.0}};
        \node[right=0.1cm of tensor_c] (a) {\scalebox{1.75}{Tensor}};
        
        \node[below=1cm of tensor_c] (tensor_d) {\drawCube{0.1}{2}{2}{yellow!20}{1.0}};
        \node[right=0.1cm of tensor_d] (b) {\scalebox{1.75}{Gradient sync.}};
    \end{tikzpicture}
}


\stdDDP}
    \vspace{4ex}
    \end{minipage}%
    \begin{minipage}{.5\linewidth}
    \centering
    \resizebox{!}{100px}{\hspace{-125pt}\definecolor{patch0}{RGB}{127,201,127}
\definecolor{patch1}{RGB}{190,174,212}
\definecolor{patch2}{RGB}{253,192,134}
\definecolor{patch3}{RGB}{255,255,153}

\definecolor{patch4}{RGB}{56,108,176}
\definecolor{patch5}{RGB}{240,2,127}
\definecolor{patch6}{RGB}{191,91,23}
\definecolor{patch7}{RGB}{158,1,66}

\definecolor{patch8}{RGB}{228,26,28}
\definecolor{patch9}{RGB}{55,126,184}
\definecolor{patch10}{RGB}{77,175,74}
\definecolor{patch11}{RGB}{152,78,163}

\definecolor{patch12}{RGB}{255,127,0}
\definecolor{patch13}{RGB}{255,255,51}
\definecolor{patch14}{RGB}{166,86,40}
\definecolor{patch15}{RGB}{247,129,191}

\newcommand{\drawCube}[5]{
    \begin{tikzpicture}[line join=round]
      \pgfmathsetmacro{\cubex}{#1}
      \pgfmathsetmacro{\cubey}{#2}
      \pgfmathsetmacro{\cubez}{#3}
      \draw[draw=black, fill=#4, opacity=1] (0,0,0)-- ++(-\cubex,0,0) -- ++(0,-\cubey,0) -- ++(\cubex,0,0) -- cycle
       (0,0,0) -- ++(0,0,-\cubez) -- ++(0,-\cubey,0) -- ++(0,0,\cubez) -- cycle
       (0,0,0) -- ++(-\cubex,0,0) -- ++(0,0,-\cubez) -- ++(\cubex,0,0) -- cycle;
    \end{tikzpicture}
}

\newcommand{\drawCubeDiv}[5]{
    \begin{tikzpicture}[line join=round]
      \pgfmathsetmacro{\cubex}{#1}
      \pgfmathsetmacro{\cubey}{#2}
      \pgfmathsetmacro{\cubez}{#3}
      \draw[draw=black, opacity=1] (0,0,0)-- ++(-\cubex,0,0) -- ++(0,-\cubey,0) -- ++(\cubex,0,0) -- cycle
       (0,0,0) -- ++(0,0,-\cubez) -- ++(0,-\cubey,0) -- ++(0,0,\cubez) -- cycle
       (0,0,0) -- ++(-\cubex,0,0) -- ++(0,0,-\cubez) -- ++(\cubex,0,0) -- cycle;
       
       \draw[draw=black, opacity=1] (0,0,0)-- ++(-\cubex,0,0) -- ++(0,-\cubey,0) -- ++(\cubex,0,0) -- cycle
       (0,0,0) -- ++(0,0,-1) -- ++(0,-\cubey,0) -- ++(0,0,1) -- cycle
       (0,0,0) -- ++(-\cubex,0,0) -- ++(0,0,-1) -- ++(\cubex,0,0) -- cycle;
    \end{tikzpicture}
}

\newcommand{\drawCubeEmpty}[5]{
    \begin{tikzpicture}[line join=round]
      \pgfmathsetmacro{\cubex}{#1}
      \pgfmathsetmacro{\cubey}{#2}
      \pgfmathsetmacro{\cubez}{#3}
      \draw[draw=white, opacity=1] (0,0,0)-- ++(-\cubex,0,0) -- ++(0,-\cubey,0) -- ++(\cubex,0,0) -- cycle
       (0,0,0) -- ++(0,0,-\cubez) -- ++(0,-\cubey,0) -- ++(0,0,\cubez) -- cycle
       (0,0,0) -- ++(-\cubex,0,0) -- ++(0,0,-\cubez) -- ++(\cubex,0,0) -- cycle;
    \end{tikzpicture}
}

\tikzset{
    cross/.pic = {
    \draw[rotate = 45] (-#1,0) -- (#1,0);
    \draw[rotate = 45] (0,-#1) -- (0, #1);
    }
}

\newcommand{\stdDDP}{
    \begin{tikzpicture}
        \node[label={[shift={(-2.8,-2.2)}]\scalebox{2.5}{GPU-4}}] (tensor_a) at (0, 0, 0) {\drawCube{2}{2}{2}{red!30}{1.0}};
        \node (tensor_b) at (1, 0, 0) {\drawCube{0.1}{2}{2}{yellow!20}{1.0}};
        \node (tensor_b) at (2.05, 0, 0) {\drawCube{2}{2}{2}{red!30}{1.0}};
        \node (tensor_b) at (3.1, 0, 0) {\drawCube{0.1}{2}{2}{yellow!20}{1.0}};
        
        \node[label={[shift={(-2.8,-2.2)}]\scalebox{2.5}{GPU-3}}] (tensor_a1) at (0, 2, 0) {\drawCube{2}{2}{2}{red!30}{1.0}};
        \node (tensor_b) at (1, 2, 0) {\drawCube{0.1}{2}{2}{yellow!20}{1.0}};
        \node (tensor_b) at (2.05, 2, 0) {\drawCube{2}{2}{2}{red!30}{1.0}};
        \node (tensor_b) at (3.1, 2, 0) {\drawCube{0.1}{2}{2}{yellow!20}{1.0}};
        
        \node[label={[shift={(-2.8,-2.2)}]\scalebox{2.5}{GPU-2}}] (tensor_a2) at (0, 4, 0) {\drawCube{2}{2}{2}{red!30}{1.0}};
        \node (tensor_b) at (1, 4, 0) {\drawCube{0.1}{2}{2}{yellow!20}{1.0}};
        \node (tensor_b) at (2.05, 4, 0) {\drawCube{2}{2}{2}{red!30}{1.0}};
        \node (tensor_b) at (3.1, 4, 0) {\drawCube{0.1}{2}{2}{yellow!20}{1.0}};
        
        \node[label={[shift={(-2.8,-2.2)}]\scalebox{2.5}{GPU-1}}] (tensor_a3) at (0, 6, 0) {\drawCube{2}{2}{2}{red!30}{1.0}};
        \node (tensor_b) at (1, 6, 0) {\drawCube{0.1}{2}{2}{yellow!20}{1.0}};
        \node (tensor_b) at (2.05, 6, 0) {\drawCube{2}{2}{2}{red!30}{1.0}};
        \node (tensor_c) at (3.1, 6, 0) {\drawCube{0.1}{2}{2}{yellow!20}{1.0}};
        
        \draw[->, line width=0.5mm] (-1.4, -1.4) -- (4, -1.4);
        \node at (3.5, -1.9)  {\scalebox{2.5}{time}};
        
        \node[above left=0.75cm and -5cm of tensor_a3, align=center] (b){\scalebox{3}{\textcolor{stdSampler}{\bfseries SSc-FBS}}};
    
    \end{tikzpicture}
}

\newcommand{\mscvbswcDDP}{
    \begin{tikzpicture}
        \node[label={[shift={(-6.25,-1.8)}]\scalebox{2.5}{GPU-4}}] (tensor_a) at (0, 0, 0) {};
        \node at (0, 0, 3) {\drawCube{7}{0.5}{0.5}{black!10!green}{1.0}}; 
        \node[label={[shift={(-6.25,0.4)}]\scalebox{2.5}{GPU-3}}] (tensor_b) at (0, 0, 0) {};
        \node at (-1.15, 2, 2.5) {\drawCube{5}{0.75}{0.75}{black!10!green}{1.0}}; 
        \node[label={[shift={(-6.25,2.6)}]\scalebox{2.5}{GPU-2}}] (tensor_c) at (0, 0, 0) {};
        \node at (-2.4, 4.01, 1.5) {\drawCube{3}{1.25}{1.25}{black!10!green}{1.0}}; 
        \node[label={[shift={(-6.25,3.8)}]\scalebox{2.5}{GPU-1}}] (tensor_d) at (0, 0, 0) {};
        \node at (-1, 6, 4) {\drawCube{4}{1}{1}{black!10!green}{1.0}}; 
        
        \node at (3.2-1.5+1.1, -0.2, 2) {\drawCube{0.1}{2}{2}{yellow!20}{1.0}};
        \node at (3.2-1.5+1.1, 1.8, 2) {\drawCube{0.1}{2}{2}{yellow!20}{1.0}};
        \node at (3.2-1.5+1.1, 3.8, 2) {\drawCube{0.1}{2}{2}{yellow!20}{1.0}};
        \node (marked) at (3.2-1.5+1.1, 5.8, 2) {\drawCube{0.1}{2}{2}{yellow!20}{1.0}};
        
        \node at (5.40-1.95+3.4-2.3, 0.5, 2.5) {\drawCube{3}{1}{2}{black!10!green}{1.0}};
        \node at (4.44-1.95+3.4-2.3, 2.02, 1.6) {\drawCube{1}{2}{4}{black!10!green}{1.0}};
        \node at (5.42-1.95+3.4-2.3, 4.01, 3.05) {\drawCube{3}{2}{1}{black!10!green}{1.0}};
        \node at (5.45-1.95+3.4-2.3, 6.55, 3.95) {\drawCube{2}{2}{2}{black!10!green}{1.0}};
        
        \node at (8.2-2.3, -0.2, 2) {\drawCube{0.1}{2}{2}{yellow!20}{1.0}};
        \node at (8.2-2.3, 1.8, 2) {\drawCube{0.1}{2}{2}{yellow!20}{1.0}};
        \node at (8.2-2.3, 3.8, 2) {\drawCube{0.1}{2}{2}{yellow!20}{1.0}};
        \node at (8.2-2.3, 5.8, 2) {\drawCube{0.1}{2}{2}{yellow!20}{1.0}};

        \draw[->, line width=0.75mm] (-4.75, -2.35) -- (7, -2.35);
        
        \node at (6.5, -2.75)  {\scalebox{2.5}{time}};
        \node[above left=0.75cm and -3cm of marked, align=center] (b) {\scalebox{3}{\textcolor{mscvbswcSampler}{\bfseries MSc-VBSWC}}};
    
    \end{tikzpicture}
}

\newcommand{\mscDDP}{
    \begin{tikzpicture}
    
        \node[label={[shift={(-3.6,-1.75)}]\scalebox{2.5}{GPU-4}}] (tensor_a) at (0.92, 0, 3) {\drawCube{4}{2}{1}{blue!30}{1.0}};
        \node[label={[shift={(-2.85,-2.25)}]\scalebox{2.5}{GPU-3}}] at (-0.08, 2, 2.5) {\drawCube{2}{2}{2}{blue!30}{1.0}};
        \node[label={[shift={(-2.85,-3.0)}]\scalebox{2.5}{GPU-2}}] at (-0.57, 4.01, 1.5) {\drawCube{1}{2}{4}{blue!30}{1.0}};
        \node[label={[shift={(-3.85,-1.0)}]\scalebox{2.5}{GPU-1}}] at (1.5, 6.1, 4) {\drawCube{4}{1}{2}{blue!30}{1.0}};
        
        \node at (3.2-0.5, -0.2, 2) {\drawCube{0.1}{2}{2}{yellow!20}{1.0}};
        \node at (3.2-0.5, 1.8, 2) {\drawCube{0.1}{2}{2}{yellow!20}{1.0}};
        \node at (3.2-0.5, 3.8, 2) {\drawCube{0.1}{2}{2}{yellow!20}{1.0}};
        \node (marked) at (3.2-0.5, 5.8, 2) {\drawCube{0.1}{2}{2}{yellow!20}{1.0}};
        
        \node at (5.90-0.95, 0.5, 2.5) {\drawCube{4}{1}{2}{blue!30}{1.0}};
        \node at (4.44-0.95, 2.02, 1.6) {\drawCube{1}{2}{4}{blue!30}{1.0}};
        \node at (5.92-0.95, 4.01, 3.05) {\drawCube{4}{2}{1}{blue!30}{1.0}};
        \node at (5.45-0.95, 6.55, 3.95) {\drawCube{2}{2}{2}{blue!30}{1.0}};
        
        \node at (8.2-1.4, -0.2, 2) {\drawCube{0.1}{2}{2}{yellow!20}{1.0}};
        \node at (8.2-1.4, 1.8, 2) {\drawCube{0.1}{2}{2}{yellow!20}{1.0}};
        \node at (8.2-1.4, 3.8, 2) {\drawCube{0.1}{2}{2}{yellow!20}{1.0}};
        \node at (8.2-1.4, 5.8, 2) {\drawCube{0.1}{2}{2}{yellow!20}{1.0}};
        
        \draw[->, line width=0.5mm] (-2.45, -2.35) -- (7, -2.35);
        
        \node at (6.5, -2.75)  {\scalebox{2.5}{time}};
        \node[above=0.75cm of marked, align=center] (b) {\scalebox{3}{\textcolor{mscSampler}{\bfseries MSc-VBS}}};
    
    \end{tikzpicture}
}

\newcommand{\tensorNotations}{
    \begin{tikzpicture}
        \node[label={[shift={(-1.75,-2.5)}]\rotatebox{90}{ \scalebox{1.35}{Height}}}, label={[shift={(-1.2,-0.9)}]\rotatebox{50}{\scalebox{1.35}{Width}}}, label={[shift={(-0.5,-3.5)}] \scalebox{1.35}{Batch}}] (tensor_c) {\drawCube{2}{2}{2}{gray!40}{1.0}};
        \node[right=0.1cm of tensor_c] (a) {\scalebox{1.75}{Tensor}};
        
        \node[below=1cm of tensor_c] (tensor_d) {\drawCube{0.1}{2}{2}{yellow!20}{1.0}};
        \node[right=0.1cm of tensor_d] (b) {\scalebox{1.75}{Gradient sync.}};
    \end{tikzpicture}
}


\mscDDP}
    \vspace{4ex}
    \end{minipage}
    \begin{minipage}{.5\linewidth}
    \centering
    \resizebox{!}{100px}{\hspace{-27pt}\definecolor{patch0}{RGB}{127,201,127}
\definecolor{patch1}{RGB}{190,174,212}
\definecolor{patch2}{RGB}{253,192,134}
\definecolor{patch3}{RGB}{255,255,153}

\definecolor{patch4}{RGB}{56,108,176}
\definecolor{patch5}{RGB}{240,2,127}
\definecolor{patch6}{RGB}{191,91,23}
\definecolor{patch7}{RGB}{158,1,66}

\definecolor{patch8}{RGB}{228,26,28}
\definecolor{patch9}{RGB}{55,126,184}
\definecolor{patch10}{RGB}{77,175,74}
\definecolor{patch11}{RGB}{152,78,163}

\definecolor{patch12}{RGB}{255,127,0}
\definecolor{patch13}{RGB}{255,255,51}
\definecolor{patch14}{RGB}{166,86,40}
\definecolor{patch15}{RGB}{247,129,191}

\newcommand{\drawCube}[5]{
    \begin{tikzpicture}[line join=round]
      \pgfmathsetmacro{\cubex}{#1}
      \pgfmathsetmacro{\cubey}{#2}
      \pgfmathsetmacro{\cubez}{#3}
      \draw[draw=black, fill=#4, opacity=1] (0,0,0)-- ++(-\cubex,0,0) -- ++(0,-\cubey,0) -- ++(\cubex,0,0) -- cycle
       (0,0,0) -- ++(0,0,-\cubez) -- ++(0,-\cubey,0) -- ++(0,0,\cubez) -- cycle
       (0,0,0) -- ++(-\cubex,0,0) -- ++(0,0,-\cubez) -- ++(\cubex,0,0) -- cycle;
    \end{tikzpicture}
}

\newcommand{\drawCubeDiv}[5]{
    \begin{tikzpicture}[line join=round]
      \pgfmathsetmacro{\cubex}{#1}
      \pgfmathsetmacro{\cubey}{#2}
      \pgfmathsetmacro{\cubez}{#3}
      \draw[draw=black, opacity=1] (0,0,0)-- ++(-\cubex,0,0) -- ++(0,-\cubey,0) -- ++(\cubex,0,0) -- cycle
       (0,0,0) -- ++(0,0,-\cubez) -- ++(0,-\cubey,0) -- ++(0,0,\cubez) -- cycle
       (0,0,0) -- ++(-\cubex,0,0) -- ++(0,0,-\cubez) -- ++(\cubex,0,0) -- cycle;
       
       \draw[draw=black, opacity=1] (0,0,0)-- ++(-\cubex,0,0) -- ++(0,-\cubey,0) -- ++(\cubex,0,0) -- cycle
       (0,0,0) -- ++(0,0,-1) -- ++(0,-\cubey,0) -- ++(0,0,1) -- cycle
       (0,0,0) -- ++(-\cubex,0,0) -- ++(0,0,-1) -- ++(\cubex,0,0) -- cycle;
    \end{tikzpicture}
}

\newcommand{\drawCubeEmpty}[5]{
    \begin{tikzpicture}[line join=round]
      \pgfmathsetmacro{\cubex}{#1}
      \pgfmathsetmacro{\cubey}{#2}
      \pgfmathsetmacro{\cubez}{#3}
      \draw[draw=white, opacity=1] (0,0,0)-- ++(-\cubex,0,0) -- ++(0,-\cubey,0) -- ++(\cubex,0,0) -- cycle
       (0,0,0) -- ++(0,0,-\cubez) -- ++(0,-\cubey,0) -- ++(0,0,\cubez) -- cycle
       (0,0,0) -- ++(-\cubex,0,0) -- ++(0,0,-\cubez) -- ++(\cubex,0,0) -- cycle;
    \end{tikzpicture}
}

\tikzset{
    cross/.pic = {
    \draw[rotate = 45] (-#1,0) -- (#1,0);
    \draw[rotate = 45] (0,-#1) -- (0, #1);
    }
}

\newcommand{\stdDDP}{
    \begin{tikzpicture}
        \node[label={[shift={(-2.8,-2.2)}]\scalebox{2.5}{GPU-4}}] (tensor_a) at (0, 0, 0) {\drawCube{2}{2}{2}{red!30}{1.0}};
        \node (tensor_b) at (1, 0, 0) {\drawCube{0.1}{2}{2}{yellow!20}{1.0}};
        \node (tensor_b) at (2.05, 0, 0) {\drawCube{2}{2}{2}{red!30}{1.0}};
        \node (tensor_b) at (3.1, 0, 0) {\drawCube{0.1}{2}{2}{yellow!20}{1.0}};
        
        \node[label={[shift={(-2.8,-2.2)}]\scalebox{2.5}{GPU-3}}] (tensor_a1) at (0, 2, 0) {\drawCube{2}{2}{2}{red!30}{1.0}};
        \node (tensor_b) at (1, 2, 0) {\drawCube{0.1}{2}{2}{yellow!20}{1.0}};
        \node (tensor_b) at (2.05, 2, 0) {\drawCube{2}{2}{2}{red!30}{1.0}};
        \node (tensor_b) at (3.1, 2, 0) {\drawCube{0.1}{2}{2}{yellow!20}{1.0}};
        
        \node[label={[shift={(-2.8,-2.2)}]\scalebox{2.5}{GPU-2}}] (tensor_a2) at (0, 4, 0) {\drawCube{2}{2}{2}{red!30}{1.0}};
        \node (tensor_b) at (1, 4, 0) {\drawCube{0.1}{2}{2}{yellow!20}{1.0}};
        \node (tensor_b) at (2.05, 4, 0) {\drawCube{2}{2}{2}{red!30}{1.0}};
        \node (tensor_b) at (3.1, 4, 0) {\drawCube{0.1}{2}{2}{yellow!20}{1.0}};
        
        \node[label={[shift={(-2.8,-2.2)}]\scalebox{2.5}{GPU-1}}] (tensor_a3) at (0, 6, 0) {\drawCube{2}{2}{2}{red!30}{1.0}};
        \node (tensor_b) at (1, 6, 0) {\drawCube{0.1}{2}{2}{yellow!20}{1.0}};
        \node (tensor_b) at (2.05, 6, 0) {\drawCube{2}{2}{2}{red!30}{1.0}};
        \node (tensor_c) at (3.1, 6, 0) {\drawCube{0.1}{2}{2}{yellow!20}{1.0}};
        
        \draw[->, line width=0.5mm] (-1.4, -1.4) -- (4, -1.4);
        \node at (3.5, -1.9)  {\scalebox{2.5}{time}};
        
        \node[above left=0.75cm and -5cm of tensor_a3, align=center] (b){\scalebox{3}{\textcolor{stdSampler}{\bfseries SSc-FBS}}};
    
    \end{tikzpicture}
}

\newcommand{\mscvbswcDDP}{
    \begin{tikzpicture}
        \node[label={[shift={(-6.25,-1.8)}]\scalebox{2.5}{GPU-4}}] (tensor_a) at (0, 0, 0) {};
        \node at (0, 0, 3) {\drawCube{7}{0.5}{0.5}{black!10!green}{1.0}}; 
        \node[label={[shift={(-6.25,0.4)}]\scalebox{2.5}{GPU-3}}] (tensor_b) at (0, 0, 0) {};
        \node at (-1.15, 2, 2.5) {\drawCube{5}{0.75}{0.75}{black!10!green}{1.0}}; 
        \node[label={[shift={(-6.25,2.6)}]\scalebox{2.5}{GPU-2}}] (tensor_c) at (0, 0, 0) {};
        \node at (-2.4, 4.01, 1.5) {\drawCube{3}{1.25}{1.25}{black!10!green}{1.0}}; 
        \node[label={[shift={(-6.25,3.8)}]\scalebox{2.5}{GPU-1}}] (tensor_d) at (0, 0, 0) {};
        \node at (-1, 6, 4) {\drawCube{4}{1}{1}{black!10!green}{1.0}}; 
        
        \node at (3.2-1.5+1.1, -0.2, 2) {\drawCube{0.1}{2}{2}{yellow!20}{1.0}};
        \node at (3.2-1.5+1.1, 1.8, 2) {\drawCube{0.1}{2}{2}{yellow!20}{1.0}};
        \node at (3.2-1.5+1.1, 3.8, 2) {\drawCube{0.1}{2}{2}{yellow!20}{1.0}};
        \node (marked) at (3.2-1.5+1.1, 5.8, 2) {\drawCube{0.1}{2}{2}{yellow!20}{1.0}};
        
        \node at (5.40-1.95+3.4-2.3, 0.5, 2.5) {\drawCube{3}{1}{2}{black!10!green}{1.0}};
        \node at (4.44-1.95+3.4-2.3, 2.02, 1.6) {\drawCube{1}{2}{4}{black!10!green}{1.0}};
        \node at (5.42-1.95+3.4-2.3, 4.01, 3.05) {\drawCube{3}{2}{1}{black!10!green}{1.0}};
        \node at (5.45-1.95+3.4-2.3, 6.55, 3.95) {\drawCube{2}{2}{2}{black!10!green}{1.0}};
        
        \node at (8.2-2.3, -0.2, 2) {\drawCube{0.1}{2}{2}{yellow!20}{1.0}};
        \node at (8.2-2.3, 1.8, 2) {\drawCube{0.1}{2}{2}{yellow!20}{1.0}};
        \node at (8.2-2.3, 3.8, 2) {\drawCube{0.1}{2}{2}{yellow!20}{1.0}};
        \node at (8.2-2.3, 5.8, 2) {\drawCube{0.1}{2}{2}{yellow!20}{1.0}};

        \draw[->, line width=0.75mm] (-4.75, -2.35) -- (7, -2.35);
        
        \node at (6.5, -2.75)  {\scalebox{2.5}{time}};
        \node[above left=0.75cm and -3cm of marked, align=center] (b) {\scalebox{3}{\textcolor{mscvbswcSampler}{\bfseries MSc-VBSWC}}};
    
    \end{tikzpicture}
}

\newcommand{\mscDDP}{
    \begin{tikzpicture}
    
        \node[label={[shift={(-3.6,-1.75)}]\scalebox{2.5}{GPU-4}}] (tensor_a) at (0.92, 0, 3) {\drawCube{4}{2}{1}{blue!30}{1.0}};
        \node[label={[shift={(-2.85,-2.25)}]\scalebox{2.5}{GPU-3}}] at (-0.08, 2, 2.5) {\drawCube{2}{2}{2}{blue!30}{1.0}};
        \node[label={[shift={(-2.85,-3.0)}]\scalebox{2.5}{GPU-2}}] at (-0.57, 4.01, 1.5) {\drawCube{1}{2}{4}{blue!30}{1.0}};
        \node[label={[shift={(-3.85,-1.0)}]\scalebox{2.5}{GPU-1}}] at (1.5, 6.1, 4) {\drawCube{4}{1}{2}{blue!30}{1.0}};
        
        \node at (3.2-0.5, -0.2, 2) {\drawCube{0.1}{2}{2}{yellow!20}{1.0}};
        \node at (3.2-0.5, 1.8, 2) {\drawCube{0.1}{2}{2}{yellow!20}{1.0}};
        \node at (3.2-0.5, 3.8, 2) {\drawCube{0.1}{2}{2}{yellow!20}{1.0}};
        \node (marked) at (3.2-0.5, 5.8, 2) {\drawCube{0.1}{2}{2}{yellow!20}{1.0}};
        
        \node at (5.90-0.95, 0.5, 2.5) {\drawCube{4}{1}{2}{blue!30}{1.0}};
        \node at (4.44-0.95, 2.02, 1.6) {\drawCube{1}{2}{4}{blue!30}{1.0}};
        \node at (5.92-0.95, 4.01, 3.05) {\drawCube{4}{2}{1}{blue!30}{1.0}};
        \node at (5.45-0.95, 6.55, 3.95) {\drawCube{2}{2}{2}{blue!30}{1.0}};
        
        \node at (8.2-1.4, -0.2, 2) {\drawCube{0.1}{2}{2}{yellow!20}{1.0}};
        \node at (8.2-1.4, 1.8, 2) {\drawCube{0.1}{2}{2}{yellow!20}{1.0}};
        \node at (8.2-1.4, 3.8, 2) {\drawCube{0.1}{2}{2}{yellow!20}{1.0}};
        \node at (8.2-1.4, 5.8, 2) {\drawCube{0.1}{2}{2}{yellow!20}{1.0}};
        
        \draw[->, line width=0.5mm] (-2.45, -2.35) -- (7, -2.35);
        
        \node at (6.5, -2.75)  {\scalebox{2.5}{time}};
        \node[above=0.75cm of marked, align=center] (b) {\scalebox{3}{\textcolor{mscSampler}{\bfseries MSc-VBS}}};
    
    \end{tikzpicture}
}

\newcommand{\tensorNotations}{
    \begin{tikzpicture}
        \node[label={[shift={(-1.75,-2.5)}]\rotatebox{90}{ \scalebox{1.35}{Height}}}, label={[shift={(-1.2,-0.9)}]\rotatebox{50}{\scalebox{1.35}{Width}}}, label={[shift={(-0.5,-3.5)}] \scalebox{1.35}{Batch}}] (tensor_c) {\drawCube{2}{2}{2}{gray!40}{1.0}};
        \node[right=0.1cm of tensor_c] (a) {\scalebox{1.75}{Tensor}};
        
        \node[below=1cm of tensor_c] (tensor_d) {\drawCube{0.1}{2}{2}{yellow!20}{1.0}};
        \node[right=0.1cm of tensor_d] (b) {\scalebox{1.75}{Gradient sync.}};
    \end{tikzpicture}
}


\mscvbswcDDP}
    \end{minipage}%
    \begin{minipage}{.5\linewidth}
    \centering
    \resizebox{!}{75px}{\hspace{-10pt}\definecolor{patch0}{RGB}{127,201,127}
\definecolor{patch1}{RGB}{190,174,212}
\definecolor{patch2}{RGB}{253,192,134}
\definecolor{patch3}{RGB}{255,255,153}

\definecolor{patch4}{RGB}{56,108,176}
\definecolor{patch5}{RGB}{240,2,127}
\definecolor{patch6}{RGB}{191,91,23}
\definecolor{patch7}{RGB}{158,1,66}

\definecolor{patch8}{RGB}{228,26,28}
\definecolor{patch9}{RGB}{55,126,184}
\definecolor{patch10}{RGB}{77,175,74}
\definecolor{patch11}{RGB}{152,78,163}

\definecolor{patch12}{RGB}{255,127,0}
\definecolor{patch13}{RGB}{255,255,51}
\definecolor{patch14}{RGB}{166,86,40}
\definecolor{patch15}{RGB}{247,129,191}

\newcommand{\drawCube}[5]{
    \begin{tikzpicture}[line join=round]
      \pgfmathsetmacro{\cubex}{#1}
      \pgfmathsetmacro{\cubey}{#2}
      \pgfmathsetmacro{\cubez}{#3}
      \draw[draw=black, fill=#4, opacity=1] (0,0,0)-- ++(-\cubex,0,0) -- ++(0,-\cubey,0) -- ++(\cubex,0,0) -- cycle
       (0,0,0) -- ++(0,0,-\cubez) -- ++(0,-\cubey,0) -- ++(0,0,\cubez) -- cycle
       (0,0,0) -- ++(-\cubex,0,0) -- ++(0,0,-\cubez) -- ++(\cubex,0,0) -- cycle;
    \end{tikzpicture}
}

\newcommand{\drawCubeDiv}[5]{
    \begin{tikzpicture}[line join=round]
      \pgfmathsetmacro{\cubex}{#1}
      \pgfmathsetmacro{\cubey}{#2}
      \pgfmathsetmacro{\cubez}{#3}
      \draw[draw=black, opacity=1] (0,0,0)-- ++(-\cubex,0,0) -- ++(0,-\cubey,0) -- ++(\cubex,0,0) -- cycle
       (0,0,0) -- ++(0,0,-\cubez) -- ++(0,-\cubey,0) -- ++(0,0,\cubez) -- cycle
       (0,0,0) -- ++(-\cubex,0,0) -- ++(0,0,-\cubez) -- ++(\cubex,0,0) -- cycle;
       
       \draw[draw=black, opacity=1] (0,0,0)-- ++(-\cubex,0,0) -- ++(0,-\cubey,0) -- ++(\cubex,0,0) -- cycle
       (0,0,0) -- ++(0,0,-1) -- ++(0,-\cubey,0) -- ++(0,0,1) -- cycle
       (0,0,0) -- ++(-\cubex,0,0) -- ++(0,0,-1) -- ++(\cubex,0,0) -- cycle;
    \end{tikzpicture}
}

\newcommand{\drawCubeEmpty}[5]{
    \begin{tikzpicture}[line join=round]
      \pgfmathsetmacro{\cubex}{#1}
      \pgfmathsetmacro{\cubey}{#2}
      \pgfmathsetmacro{\cubez}{#3}
      \draw[draw=white, opacity=1] (0,0,0)-- ++(-\cubex,0,0) -- ++(0,-\cubey,0) -- ++(\cubex,0,0) -- cycle
       (0,0,0) -- ++(0,0,-\cubez) -- ++(0,-\cubey,0) -- ++(0,0,\cubez) -- cycle
       (0,0,0) -- ++(-\cubex,0,0) -- ++(0,0,-\cubez) -- ++(\cubex,0,0) -- cycle;
    \end{tikzpicture}
}

\tikzset{
    cross/.pic = {
    \draw[rotate = 45] (-#1,0) -- (#1,0);
    \draw[rotate = 45] (0,-#1) -- (0, #1);
    }
}

\newcommand{\stdDDP}{
    \begin{tikzpicture}
        \node[label={[shift={(-2.8,-2.2)}]\scalebox{2.5}{GPU-4}}] (tensor_a) at (0, 0, 0) {\drawCube{2}{2}{2}{red!30}{1.0}};
        \node (tensor_b) at (1, 0, 0) {\drawCube{0.1}{2}{2}{yellow!20}{1.0}};
        \node (tensor_b) at (2.05, 0, 0) {\drawCube{2}{2}{2}{red!30}{1.0}};
        \node (tensor_b) at (3.1, 0, 0) {\drawCube{0.1}{2}{2}{yellow!20}{1.0}};
        
        \node[label={[shift={(-2.8,-2.2)}]\scalebox{2.5}{GPU-3}}] (tensor_a1) at (0, 2, 0) {\drawCube{2}{2}{2}{red!30}{1.0}};
        \node (tensor_b) at (1, 2, 0) {\drawCube{0.1}{2}{2}{yellow!20}{1.0}};
        \node (tensor_b) at (2.05, 2, 0) {\drawCube{2}{2}{2}{red!30}{1.0}};
        \node (tensor_b) at (3.1, 2, 0) {\drawCube{0.1}{2}{2}{yellow!20}{1.0}};
        
        \node[label={[shift={(-2.8,-2.2)}]\scalebox{2.5}{GPU-2}}] (tensor_a2) at (0, 4, 0) {\drawCube{2}{2}{2}{red!30}{1.0}};
        \node (tensor_b) at (1, 4, 0) {\drawCube{0.1}{2}{2}{yellow!20}{1.0}};
        \node (tensor_b) at (2.05, 4, 0) {\drawCube{2}{2}{2}{red!30}{1.0}};
        \node (tensor_b) at (3.1, 4, 0) {\drawCube{0.1}{2}{2}{yellow!20}{1.0}};
        
        \node[label={[shift={(-2.8,-2.2)}]\scalebox{2.5}{GPU-1}}] (tensor_a3) at (0, 6, 0) {\drawCube{2}{2}{2}{red!30}{1.0}};
        \node (tensor_b) at (1, 6, 0) {\drawCube{0.1}{2}{2}{yellow!20}{1.0}};
        \node (tensor_b) at (2.05, 6, 0) {\drawCube{2}{2}{2}{red!30}{1.0}};
        \node (tensor_c) at (3.1, 6, 0) {\drawCube{0.1}{2}{2}{yellow!20}{1.0}};
        
        \draw[->, line width=0.5mm] (-1.4, -1.4) -- (4, -1.4);
        \node at (3.5, -1.9)  {\scalebox{2.5}{time}};
        
        \node[above left=0.75cm and -5cm of tensor_a3, align=center] (b){\scalebox{3}{\textcolor{stdSampler}{\bfseries SSc-FBS}}};
    
    \end{tikzpicture}
}

\newcommand{\mscvbswcDDP}{
    \begin{tikzpicture}
        \node[label={[shift={(-6.25,-1.8)}]\scalebox{2.5}{GPU-4}}] (tensor_a) at (0, 0, 0) {};
        \node at (0, 0, 3) {\drawCube{7}{0.5}{0.5}{black!10!green}{1.0}}; 
        \node[label={[shift={(-6.25,0.4)}]\scalebox{2.5}{GPU-3}}] (tensor_b) at (0, 0, 0) {};
        \node at (-1.15, 2, 2.5) {\drawCube{5}{0.75}{0.75}{black!10!green}{1.0}}; 
        \node[label={[shift={(-6.25,2.6)}]\scalebox{2.5}{GPU-2}}] (tensor_c) at (0, 0, 0) {};
        \node at (-2.4, 4.01, 1.5) {\drawCube{3}{1.25}{1.25}{black!10!green}{1.0}}; 
        \node[label={[shift={(-6.25,3.8)}]\scalebox{2.5}{GPU-1}}] (tensor_d) at (0, 0, 0) {};
        \node at (-1, 6, 4) {\drawCube{4}{1}{1}{black!10!green}{1.0}}; 
        
        \node at (3.2-1.5+1.1, -0.2, 2) {\drawCube{0.1}{2}{2}{yellow!20}{1.0}};
        \node at (3.2-1.5+1.1, 1.8, 2) {\drawCube{0.1}{2}{2}{yellow!20}{1.0}};
        \node at (3.2-1.5+1.1, 3.8, 2) {\drawCube{0.1}{2}{2}{yellow!20}{1.0}};
        \node (marked) at (3.2-1.5+1.1, 5.8, 2) {\drawCube{0.1}{2}{2}{yellow!20}{1.0}};
        
        \node at (5.40-1.95+3.4-2.3, 0.5, 2.5) {\drawCube{3}{1}{2}{black!10!green}{1.0}};
        \node at (4.44-1.95+3.4-2.3, 2.02, 1.6) {\drawCube{1}{2}{4}{black!10!green}{1.0}};
        \node at (5.42-1.95+3.4-2.3, 4.01, 3.05) {\drawCube{3}{2}{1}{black!10!green}{1.0}};
        \node at (5.45-1.95+3.4-2.3, 6.55, 3.95) {\drawCube{2}{2}{2}{black!10!green}{1.0}};
        
        \node at (8.2-2.3, -0.2, 2) {\drawCube{0.1}{2}{2}{yellow!20}{1.0}};
        \node at (8.2-2.3, 1.8, 2) {\drawCube{0.1}{2}{2}{yellow!20}{1.0}};
        \node at (8.2-2.3, 3.8, 2) {\drawCube{0.1}{2}{2}{yellow!20}{1.0}};
        \node at (8.2-2.3, 5.8, 2) {\drawCube{0.1}{2}{2}{yellow!20}{1.0}};

        \draw[->, line width=0.75mm] (-4.75, -2.35) -- (7, -2.35);
        
        \node at (6.5, -2.75)  {\scalebox{2.5}{time}};
        \node[above left=0.75cm and -3cm of marked, align=center] (b) {\scalebox{3}{\textcolor{mscvbswcSampler}{\bfseries MSc-VBSWC}}};
    
    \end{tikzpicture}
}

\newcommand{\mscDDP}{
    \begin{tikzpicture}
    
        \node[label={[shift={(-3.6,-1.75)}]\scalebox{2.5}{GPU-4}}] (tensor_a) at (0.92, 0, 3) {\drawCube{4}{2}{1}{blue!30}{1.0}};
        \node[label={[shift={(-2.85,-2.25)}]\scalebox{2.5}{GPU-3}}] at (-0.08, 2, 2.5) {\drawCube{2}{2}{2}{blue!30}{1.0}};
        \node[label={[shift={(-2.85,-3.0)}]\scalebox{2.5}{GPU-2}}] at (-0.57, 4.01, 1.5) {\drawCube{1}{2}{4}{blue!30}{1.0}};
        \node[label={[shift={(-3.85,-1.0)}]\scalebox{2.5}{GPU-1}}] at (1.5, 6.1, 4) {\drawCube{4}{1}{2}{blue!30}{1.0}};
        
        \node at (3.2-0.5, -0.2, 2) {\drawCube{0.1}{2}{2}{yellow!20}{1.0}};
        \node at (3.2-0.5, 1.8, 2) {\drawCube{0.1}{2}{2}{yellow!20}{1.0}};
        \node at (3.2-0.5, 3.8, 2) {\drawCube{0.1}{2}{2}{yellow!20}{1.0}};
        \node (marked) at (3.2-0.5, 5.8, 2) {\drawCube{0.1}{2}{2}{yellow!20}{1.0}};
        
        \node at (5.90-0.95, 0.5, 2.5) {\drawCube{4}{1}{2}{blue!30}{1.0}};
        \node at (4.44-0.95, 2.02, 1.6) {\drawCube{1}{2}{4}{blue!30}{1.0}};
        \node at (5.92-0.95, 4.01, 3.05) {\drawCube{4}{2}{1}{blue!30}{1.0}};
        \node at (5.45-0.95, 6.55, 3.95) {\drawCube{2}{2}{2}{blue!30}{1.0}};
        
        \node at (8.2-1.4, -0.2, 2) {\drawCube{0.1}{2}{2}{yellow!20}{1.0}};
        \node at (8.2-1.4, 1.8, 2) {\drawCube{0.1}{2}{2}{yellow!20}{1.0}};
        \node at (8.2-1.4, 3.8, 2) {\drawCube{0.1}{2}{2}{yellow!20}{1.0}};
        \node at (8.2-1.4, 5.8, 2) {\drawCube{0.1}{2}{2}{yellow!20}{1.0}};
        
        \draw[->, line width=0.5mm] (-2.45, -2.35) -- (7, -2.35);
        
        \node at (6.5, -2.75)  {\scalebox{2.5}{time}};
        \node[above=0.75cm of marked, align=center] (b) {\scalebox{3}{\textcolor{mscSampler}{\bfseries MSc-VBS}}};
    
    \end{tikzpicture}
}

\newcommand{\tensorNotations}{
    \begin{tikzpicture}
        \node[label={[shift={(-1.75,-2.5)}]\rotatebox{90}{ \scalebox{1.35}{Height}}}, label={[shift={(-1.2,-0.9)}]\rotatebox{50}{\scalebox{1.35}{Width}}}, label={[shift={(-0.5,-3.5)}] \scalebox{1.35}{Batch}}] (tensor_c) {\drawCube{2}{2}{2}{gray!40}{1.0}};
        \node[right=0.1cm of tensor_c] (a) {\scalebox{1.75}{Tensor}};
        
        \node[below=1cm of tensor_c] (tensor_d) {\drawCube{0.1}{2}{2}{yellow!20}{1.0}};
        \node[right=0.1cm of tensor_d] (b) {\scalebox{1.75}{Gradient sync.}};
    \end{tikzpicture}
}


\tensorNotations}
    \end{minipage}\par\medskip
    \caption{\textbf{Single-scale fixed batch size (SSc-FBS) vs multi-scale variable batch size (MSc-VBS) vs multi-scale variable batch size with curriculum (MSc-VBSWC) samplers.} In the SSc-FBS sampler, at each training iteration, each GPU receives a batch of data that is the same shape throughout training. In the MSc-VBS sampler, at each training iteration, each GPU will randomly sample a training resolution and dynamically adjust the batch size to use a large batch for small resolutions, and a small batch for large resolutions. In the MSc-VBSWC sampler, the sample resolutions expand throughout the course of training while leveraging the dynamic batch sizes of the MSc-VBS sampler.}
    \label{fig:sampler_illustration}
\end{figure}
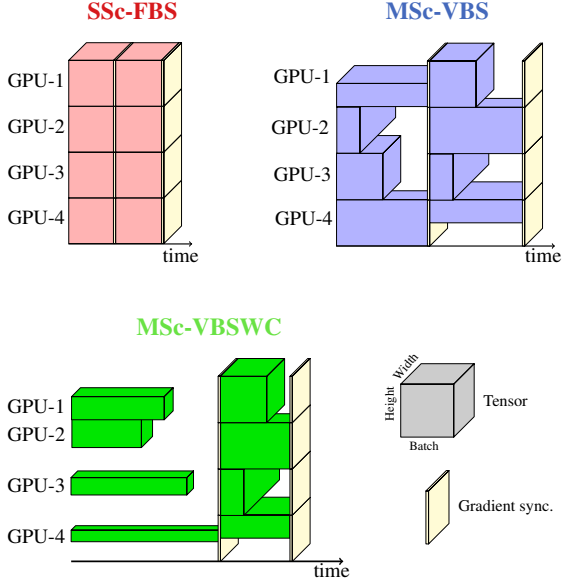

\paragraph{Single-scale fixed batch size}
The single-scale fixed batch size (SSc-FBS) data sampler is the default sampling procedure that is implemented across many deep learning libraries and frameworks \cite{timm, paszke2019pytorch, tensorflow2015-whitepaper}. In this setting, the batch shape at iteration $t$, $\gBs_t$, remains fixed so that $\gBs_t = (B, C, H, W)$ for all iterations $t$.

\paragraph{Multi-scale fixed batch size} SSc-FBS allows the model to learn representations at single scale. However, in the real world, objects appear at different scales. Multi-scale samplers with fixed batch (MSc-FBS) \cite{yolo9000, cvnets} addresses this shortcoming of SSc-FBS, and enables training of a model at multiple input scales. More specifically, MSc-FBS randomly samples a batch with shape $\gBs_t = (B, C, H_t, W_t)$ at each training iteration $t$, where $(H_t, W_t)$ is drawn randomly from a set of possible spatial resolutions $\gS=\{(H_1, W_1), (H_2, W_2), \ldots, (H_n, W_n)\}$ .

\paragraph{Multi-scale with variable batch size}  One drawback of MSc-FBS is its fixed batch size selection. Because the batch size $B$ is fixed, if a large resolution $(H_t,W_t) \in \gS$ is chosen, an out-of-memory (OOM) error can occur. Thus, the choices of $(H_t,W_t)$ are limited by $B$. Conversely, a small spatial resolution may lead to under-utilization of computational resources. To address this, multi-scale variable batch size samplers (MSc-VBS) were introduced in \cite{MobileViT, cvnets}. This sampler circumvents the compute bottleneck of the MSc-FBS sampler by dynamically adjusting the batch size $B$ according to the resolution $(H_t, W_t)$. In particular, a ``reference'' batch shape $(B, C, H, W)$ is first defined. Then, at training iteration $t$, the MSc-VBS samples a batch with shape $\gBs_t = (B_t, C, H_t, W_t)$ where $(H_t, W_t) \in \gS$ are sampled randomly and $B_t = \frac{HW}{H_tW_t}B$. Thus, MSc-VBS will use a larger batch size $B_t$ when the resolution $(H_t, W_t)$ is small, and a smaller $B_t$ when the resolution is large. The MSc-VBS sampler dynamically controls the batch size to offset the change in compute due to the spatial resolution, allowing it to efficiently utilize the underlying hardware and avoid OOM errors during training. Figure \ref{fig:sampler_illustration} visualizes single-scale and multi-scale variable batch samplers.

\paragraph{Multi-scale with variable batch size and curriculum} Motivated by curriculum and progressive learning \cite{curriculum_learning, progressive-resizing-howard2018, tan2021efficientnetv2}, we extend the MSc-VBS sampler with a curriculum which yields the MSc-VBSWC sampler. We compare and contrast the properties of MSc-VBSWC against the MSc-FBS and MSc-VBS samplers. Similar to MSc-VBS, we first consider a set of possible spatial resolutions $\gS$ and a reference batch shape $(B, C, H, W)$. In this setting, $\gS$ serves as a ``reference'' pool of spatial resolutions that our sampler expands to. In particular, at each training epoch $e$, we consider a pool of sample resolutions $s(e) = \{(h_1(e), w_1(e)), \ldots, (h_n(e), w_n(e))\}$. The spatial resolutions $(h_i(e), w_i(e))$  progressively grow to $(H_i, W_i) \in \gS$ for all $i=1, \ldots, n$ over the course of training. Given an initial compression factor $0 < \rho_0 \leq 1$ and expansion period $0 < \tau \leq 1$, in the first epoch we sample from $s(0) = \{(\rho_0\cdot H_1, \rho_0\cdot W_1), \ldots, (\rho_0\cdot H_n, \rho_0\cdot W_n)\}$ and after $\tau E$ epochs we sample from $s(\tau E) = \gS$. In words, at the start of training, we sample from a ``compressed'' form of $\gS$ where each spatial resolution is scaled by $\rho_0$, then, as training progresses, the set of spatial resolutions we sample from, $s(e)$, expands to $\gS$ over the first $\tau E$ epochs. The rate at which this expansion occurs is controlled by a schedule, such as a cosine or linear schedule. A depiction of this sampler is provided in Figure \ref{fig:sampler_illustration} and we provide further details in the Appendix, where we run ablation experiments on the choices of curriculum thresholds and found $\rho=0.75$ and $\tau=0.5$ to perform the well. We use this setup throughout the rest of the paper.

\begin{table*}[t!]
    \centering
    \begin{tabular}{lccccc}
        \toprule[1.5pt]
        \textbf{Sampler} & \textbf{Peak GPU Memory} &  \textbf{Training FLOPs} & \textbf{Optimization updates} & \textbf{Training time} & \textbf{Top-1 Accuracy} (\%)\\
        \midrule[1.25pt]
        SSc-FBS   & $\mathbf{1.00\times}$    & $1.00\times$    & $1.00\times$    & $1.00\times$    & 81.31 \\
        MSc-FBS   & $2.23\times$ & $1.15\times$ & $1.00\times$    & $1.15\times$ & 81.01 \\
        MSc-VBS   & $1.22\times$ & $0.77\times$ & $0.77\times$ & $0.92\times$ & \textbf{81.66} \\
        MSc-VBSWC & $1.75\times$ & $\mathbf{0.70\times}$ & $\mathbf{0.66\times}$ & $\mathbf{0.84\times}$ & 81.53 \\
         \bottomrule[1.5pt]
    \end{tabular}
    \caption{\textbf{Training with a multi-scale variable batch size sampler promotes faster training.} Here we train a ResNet-101 model on ImageNet with single-scale (SSc-FBS) and multi-scale (MSc-FBS, MSc-VBS, MSc-VBSWC) samplers. Multi-scale training retains the accuracy of the model trained with SSc-FBS while training faster and being more computationally efficient.}
    \label{tab:resnet101_training_speed}
\end{table*}

\section{Why Multi-scale samplers?}
\label{sec:why_msc_samplers}
In this section, we analyze the properties of the different samplers in Section \ref{sec:overview} and compare their performances.  We first establish their benefits over single-scale training and show that multi-scale variable batch size samplers train faster and lead to more accurate models (Section~\ref{ssec:faster_training}). To further understand the properties of multi-scale samplers, we study their robustness and empirically show that they are better calibrated (Section \ref{sec:robustness}). Finally, we study their regularization behavior (Section ~\ref{ssec:regularization}).

We study these properties on the ImageNet \cite{ImageNet} dataset using the ResNet-101 architecture \cite{resnet} and follow the recipes in \cite{cvnets, rangeaugment} for training and evaluating the models. For results on other architectures, see Section \ref{sec:generic}. Additionally, in the Appendix Section \ref{appendix:gpus}, we provide a discussion on the effect that the number of GPUs has on the performance of single and multi-scale samplers, as well as uncertainty measurements. 

\subsection{Faster Training}\label{ssec:faster_training}

\paragraph{Metrics:} We use the following metrics to measure the training speed: (1) \textbf{Training FLOPs} measures  computational complexity during training while being hardware-agnostic. Generally, a data sampler with fewer training FLOPs is preferable. (2) \textbf{Optimization updates} measures the number of training iterations it takes to train a model. Fewer updates is generally more efficient and favorable. (3) \textbf{Peak GPU memory} measures the maximum memory that is required for training a model. Data samplers with lower memory are preferable. (4) \textbf{Training time} measures the wall-clock time taken to train the model. Data samplers with lower training times are desirable. Note that, in these experiments, we used the same hardware for training all models, as well as the same number of data workers and GPUs. However, a careful tuning of hardware resources may vary the training time significantly. Such experiments are beyond the scope of this paper.

\paragraph{Results:} Table \ref{tab:resnet101_training_speed} compares the performance of different samplers. In general, multi-scale samplers (MSc-FBS, MSc-VBS, and MSC-VBSWC) are able to match the performance of the single-scale sampler (SSc-FBS). Multi-scale samplers with variable batch sizes (MSC-VBS and MSc-VBSWC) also reduce the training FLOPs and optimization updates significantly, resulting in faster training. This is because multi-scale variable batch samplers adjust the batch size depending on the input spatial resolution (larger batch sizes are used for smaller spatial resolutions and vice-versa). Compared to training with SSc-FBS, training with MSc-VBSWC reduces training FLOPs, optimization updates, and wall clock time by 30\%, 34\%, and 16\% respectively, in our ResNet-101 ImageNet experiments, all while maintaining or slightly increasing the accuracy. Due to the large computational requirements of the MSc-FBS sampler, we focus our multi-scale sampler analysis on the MSc-VBS and MSc-VBSWC samplers in the rest of the paper.

\subsection{Robustness}\label{sec:robustness}

We observe in Section \ref{ssec:faster_training} that multi-scale samplers can preserve or slightly improve accuracy and accelerate training. In this section, we assess the robustness of models trained with multi-scale samplers compared to those trained with single-scale samplers.

\paragraph{Metrics:} We study the robustness of models trained with multi-scale samplers in five ways. 
(1) \textbf{Accuracy on standard robustness benchmarks.} We evaluate the top-1 accuracy on three standard  benchmarks for estimating a model's robustness: ImageNet-A \cite{ImageNet-A}, ImageNet-R \cite{ImageNet-R}, and ImageNetV2 \cite{ImageNet-v2}.  
(2) \textbf{Expected calibration error.} A classification model is well-calibrated if its predicted class probabilities capture the true underlying distribution correctly. This is a desirable property as it allows users of such models to more reliably accept or reject the models' outputs. A standard metric to measure the degree of a model's miscalibration is the expected calibration error (ECE) \cite{calibration_of_nn, ece_calibration}, which measures the discrepancy between the model's confidence and the accuracy. Hence, a well-calibrated model will have a low ECE. (3) \textbf{Image embedding variance.} An image embedding is a vector representation of an image that captures its semantic information. A model that produces low-variance embeddings is robust and selective \cite{goodfellow2009measuring, shang2016understanding}. (4) \textbf{Robustness to scale changes} Generally speaking, a discriminative model should be robust to scale changes. To assess this, we evaluate our models across a broad range of input resolutions. (5) \textbf{Entropy Skewness} The entropy of a classification model's predicted class distribution reflects the model's uncertainty in its prediction. By computing the entropy for each image in the validation set at each training epoch, we can obtain empirical entropy distributions that we can monitor over the course of training. Measuring the skewness of these distributions allows us to investigate the rate at which a model becomes more certain in its predictions.

\begin{figure*}[t!]
    \centering
    \begin{subfigure}[b]{0.95\columnwidth}
        \centering
        \includegraphics[width=\columnwidth]{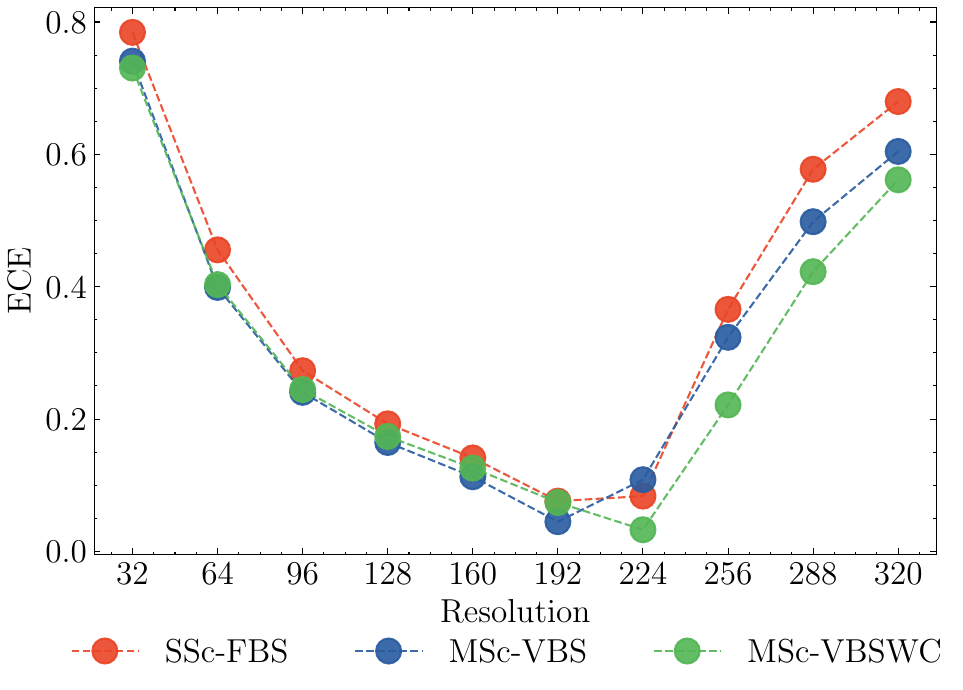}
        \caption{Calibration error}
        \label{fig:rn101_ece}
    \end{subfigure}
    \hfill
    \begin{subfigure}[b]{0.95\columnwidth}
        \centering
        \includegraphics[width=\columnwidth]{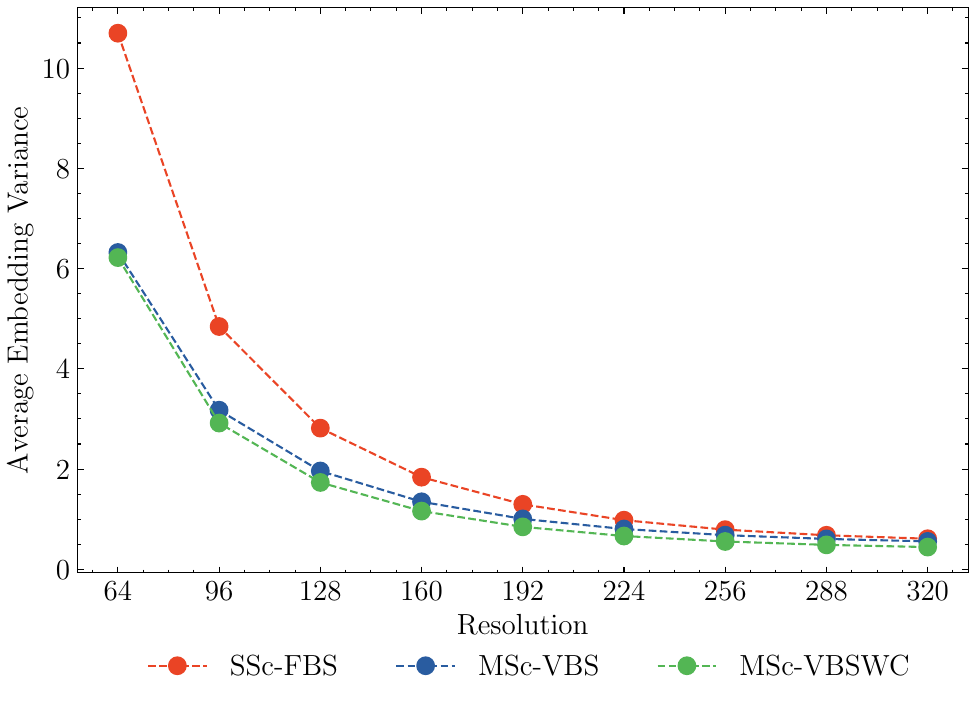}
        \caption{Embedding variance}
        \label{fig:rn101_embedding_var_vs_res}
    \end{subfigure}
    \vfill
    \begin{subfigure}[b]{0.95\columnwidth}
        \centering
        \includegraphics[width=\columnwidth]{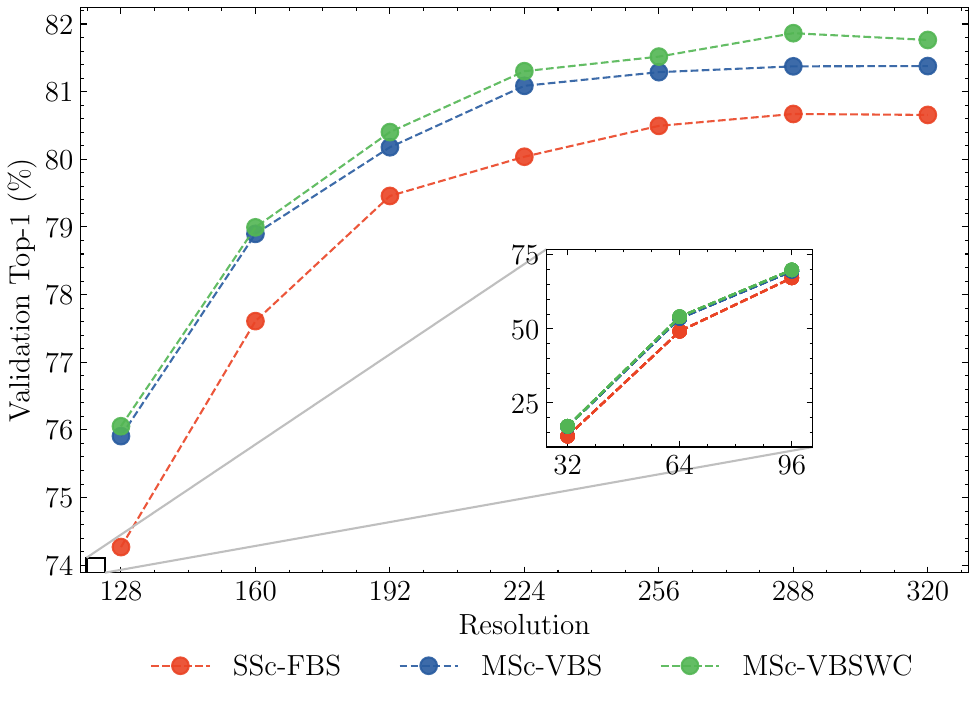}
        \caption{Accuracy across resolutions}
        \label{fig:resnet101_val_vs_resolution}
    \end{subfigure}
    \hfill
    \begin{subfigure}[b]{0.925\columnwidth}
        \centering
        \includegraphics[width=\columnwidth]{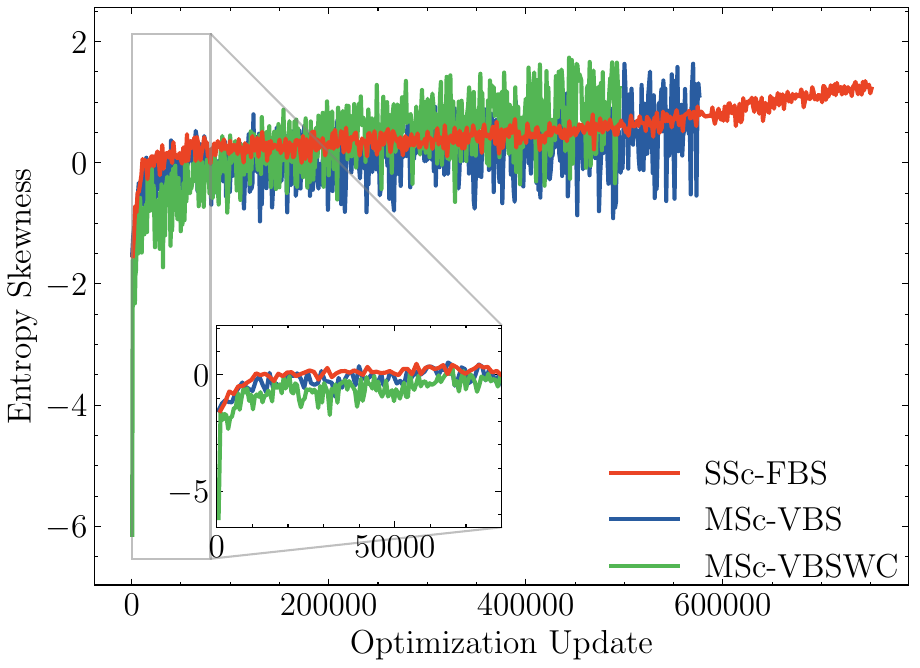}
        \caption{Skewness of empirical entropy distributions}
        \label{fig:resnet101_entropy_skewness}
    \end{subfigure}
    \caption{\textbf{Multi-scale samplers improve model calibration, reduce embedding variance, and improve robustness.} Here we train a ResNet-101 model with single-scale and multi-scale samplers. We find that models trained with multi-scale samplers are better calibrated, learn embeddings with lower variance, and are more robust to changes in input scale. Moreover, multi-scale samplers exhibit a larger shift in entropy, suggesting a larger exploration of the weight space.}
    \label{fig:robustness_plots}
\end{figure*}

\paragraph{Results:}
(1) Table \ref{table:resnet101_robustness} shows results on different robustness datasets. We observe that ResNet-101 trained with multi-scale samplers consistently achieves a higher accuracy across all benchmark datasets. Furthermore, models trained with multi-scale samplers have (2) lower ECE (Figure \ref{fig:rn101_ece}), (3) low image embedding variance (Figure \ref{fig:rn101_embedding_var_vs_res}), and (4) are less sensitive to scale changes (Figure \ref{fig:resnet101_val_vs_resolution}) as compared to single scale samplers. These results suggest that models trained with multi-scale samplers are able to more effectively extract the salient semantic information in the image (e.g., changes in image scale and rotation). Moreover,  these models are more robust and have a greater discriminative power. Our findings regarding embedding variance match \cite{johnson2013accelerating,roux2012stochastic}, which also show that a reduction in embedding variance accelerates training. (5) Additionally, the large variation in the entropy skewness of the multi-scale samplers (Figure \ref{fig:resnet101_entropy_skewness}) suggests that multi-scale samplers explore more of the weight space. Coupled with the gradual shift toward large skewness values, this suggests that the increased exploration steers the model towards more robust minima. 

\begin{table}[t!]
    \centering
    \resizebox{\columnwidth}{!}{
    \begin{tabular}{lccc}
        \toprule[1.5pt]
        \textbf{Dataset} &  \textbf{SSc-FBS} & \textbf{MSc-VBS} & \textbf{MSc-VBSWC}\\
        \midrule[1.25pt]
        ImageNet-A    & $16.05_{(0.0)}$ & $17.91_{(+1.86)}$          & $\mathbf{19.17}_{(+3.12)}$ \\
        ImageNet-R    & $39.77_{(0.0)}$ & $\mathbf{41.61}_{(+1.84)}$ & $41.57_{(+1.80)}$ \\
        ImageNetV2-MF & $68.75_{(0.0)}$ & $69.86_{(+1.11)}$          & $\mathbf{70.05}_{(+1.30)}$ \\
        ImageNetV2-Th & $76.82_{(0.0)}$ & $\mathbf{77.97}_{(+1.15)}$ & $77.66_{(+0.84)}$ \\
        ImageNetV2-TI & $81.30_{(0.0)}$ & $81.75_{(+0.45)}$          & $\mathbf{82.03}_{(+0.73)}$ \\
        \bottomrule
    \end{tabular}
    }
    \caption{\textbf{Training with multi-scale samplers improves robustness.} Here we evaluate our ResNet-101 models that were trained with single-scale and multi-scale samplers on ImageNet. We report the top-1 accuracy (\%) across multiple datasets and observe that models trained with multi-scale samplers consistently outperform the single-scale model. ImageNetV2-MF refers to the ``matched frequency'' subset of ImageNetV2, ImageNetV2-Th refers to the ``threshold 0.7'' subset, and ImageNetV2-TI refers to the ``top images'' subset (see \cite{ImageNet-v2}).}
    \label{table:resnet101_robustness}
\end{table}

\begin{figure*}[t!]
    \centering
    \begin{subfigure}[b]{0.66\columnwidth}
        \centering
        \includegraphics[width=\columnwidth]{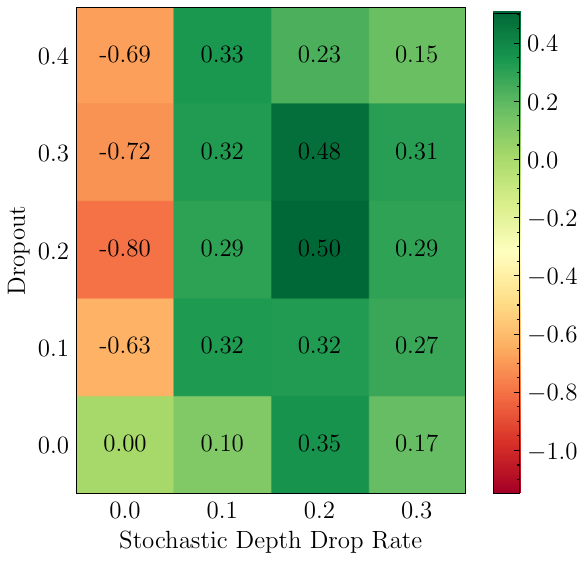}
        \caption{SSc-FBS}
        \label{fig:resnet101_cd_sd_gs_ssc_fbs}
    \end{subfigure}
    \hfill
    \begin{subfigure}[b]{0.66\columnwidth}
        \centering
        \includegraphics[width=\columnwidth]{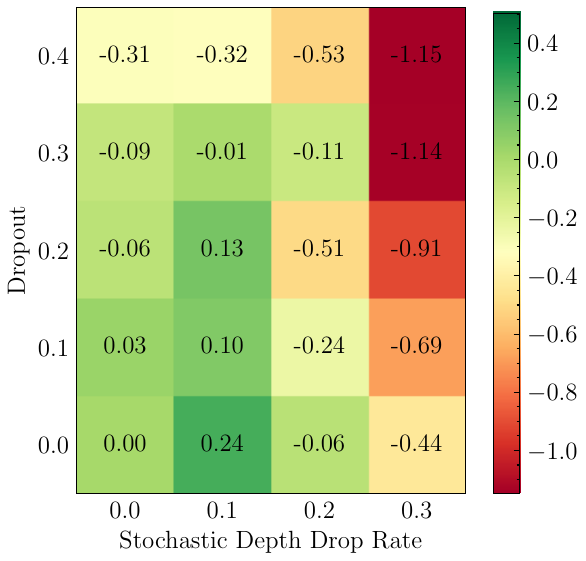}
        \caption{MSc-VBS}
        \label{fig:resnet101_cd_sd_gs_msc_vbs}
    \end{subfigure}
    \hfill
    \begin{subfigure}[b]{0.66\columnwidth}
        \centering
        \includegraphics[width=\columnwidth]{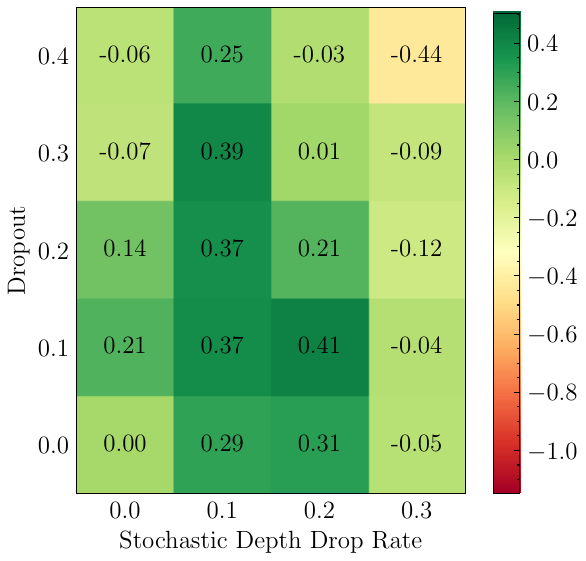}
        \caption{MSc-VBSWC}
        \label{fig:resnet101_cd_sd_gs_msc_vbswc}
    \end{subfigure}
    \caption{\textbf{Multi-scale samplers are implicit data regularizers.} We train ResNet-101 on the ImageNet dataset at different values of classifier dropout (y-axis) and stochastic depth drop rate (x-axis) with three samplers (SSc-FBS, MSc-VBS, and MSc-VBSWC). ResNet-101 trained with multi-scale samplers (MSc-VBS and MSc-VBSWC) requires less regularization as compared to SSc-FBS. Here, the values in each cell are relative to the bottom left cell. The top-1 accuracy of ResNet-101 for bottom left cell (i.e.,  the values of classifier dropout and stochastic depth are 0.0) for SSc-FBS, MSc-VBS, and MSc-VBSWC are 81.31\%, 81.66\%, and 81.53\% respectively.}
    \label{fig:resnet101_cd_sd_gs}
\end{figure*}

\subsection{Regularization}\label{ssec:regularization}
Multi-scale data augmentation and sampling methods have been adopted in previous works for improving model accuracy \cite{MobileViT}. However, to the best of our knowledge, the pairing of multi-scale data sampling with explicit regularization methods (e.g., dropout \cite{dropout}, stochastic depth drop rate \cite{stochastic_depth}, and L2 regularization) has not been extensively studied. It may be the case that multi-scale samplers offer implicit regularization, and as a consequence, require little or no explicit regularization and hence can lead to reduced parameter tuning. In this section, we study the regularization aspect of different samplers.

\paragraph{Metrics:} To assess the regularization, we consider three widely-used regularization methods in state-of-the-art models: (1) classifier dropout, (2) stochastic depth drop rate, and (3) L2 weight regularization.

\begin{figure}[t!]
    \centering
   \includegraphics[width=0.9\linewidth]{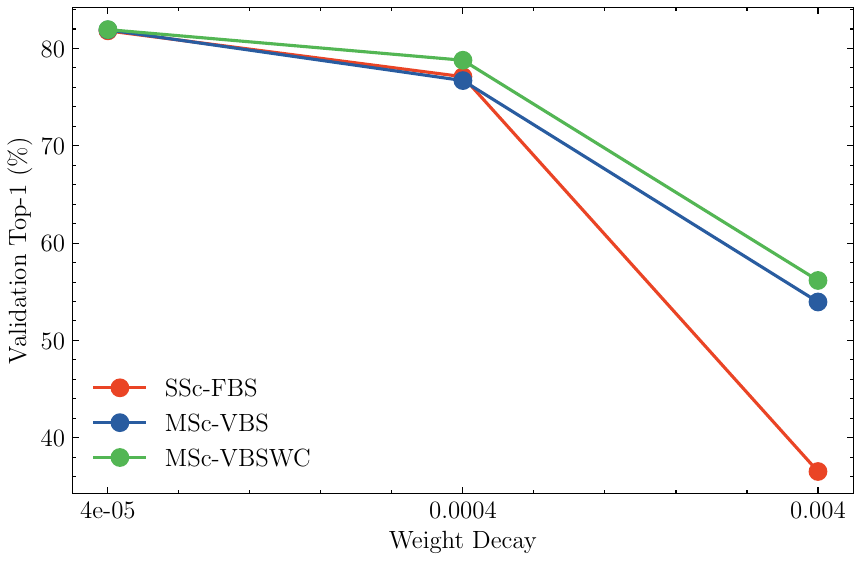}
   \caption{\textbf{Models trained with multi-scale samplers are less sensitive to weight decay strength.} Here we train ResNet-101 models on ImageNet with varying degrees of weight decay. Models trained with multi-scale samplers enjoy a lower drop in accuracy due to stronger weight decay.}
\label{fig:resnet101_wd_gs}
\end{figure}

\paragraph{Results:} Figure \ref{fig:resnet101_cd_sd_gs} shows the effect of varying the values of classifier  and stochastic dropouts during training of ResNet-101 on the ImageNet dataset. In the case of SSc-FBS training, we observe that the best top-1 accuracy of 81.81\% is obtained when strong regularization (classifier dropout: 0.2 and stochastic depth drop rate: 0.2) is used (Figure \ref{fig:resnet101_cd_sd_gs_ssc_fbs}). On the other hand, similar or better accuracy with smaller regularization coefficients is achieved when we instead train with multi-scale samplers. For example, MSc-VBS (Figure \ref{fig:resnet101_cd_sd_gs_msc_vbs}) achieves a top-1 accuracy of 81.91\% with a stochastic dropout of 0.1 (and no classifier dropout). Moreover, when there is no explicit regularization (i.e., the value of classifier and stochastic dropouts are 0.0), the multi-scale samplers increase the accuracy of ResNet-101 by 0.2-0.3\%. 

Furthermore, as shown in Figure \ref{fig:resnet101_wd_gs}, when the value of the L2 regularization coefficient is varied from $4e^{-5}$ to $4e^{-3}$, the drop in top-1 accuracy of models trained with multi-scale samplers is significantly lower than the top-1 accuracy drop of the single-scale sampler. These results, in conjunction with Figure \ref{fig:resnet101_cd_sd_gs}, suggest that ResNet-101 trained with multi-scale samplers requires less explicit regularization and is less sensitive to the choice of weight decay. This is likely because multi-scale samplers act as data dependent regularizers.

\begin{table*}[t!]
    \centering
    \resizebox{2\columnwidth}{!}{
    \begin{tabular}{llccccc}
        \toprule[1.5pt]
        \textbf{Model} & \textbf{Sampler} & \textbf{GPU Memory} &  \textbf{Training FLOPs} & \textbf{Optimization updates} & \textbf{Training time} & \textbf{Top-1 Accuracy} (\%)\\
        \midrule[1.25pt]
        ResNet-50 \cite{resnet} & SSc-FBS             & $\mathbf{1.00}\times$ & $1.00\times$       & $1.00\times$    & $1.00\times$    & 79.43 \\
                  & MSc-FBS   & $2.16\times$          & $1.15\times$ & $1.00\times$    & $1.05\times$          & 80.01 \\
                  & MSc-VBS   & $1.31\times$          & $0.77\times$ & $0.77\times$ & $0.95\times$          & 80.03 \\
                  & MSc-VBSWC & $1.03\times$ & $\mathbf{0.70}\times$ & $\mathbf{0.66}\times$ & $\mathbf{0.85}\times$ & \textbf{80.25} \\
      \midrule
        SE-ResNet-50 \cite{squeeze_excitation} & SSc-FBS   & $\mathbf{1.00}\times$ & $1.00\times$    & $1.00\times$    & $1.00\times$    & 80.24 \\
                     & MSc-FBS   & $2.21\times$ & $1.15\times$ & $1.00\times$    & $1.13\times$ & 80.64 \\
                     & MSc-VBS   & $1.32\times$ & $0.77\times$ & $0.77\times$ & $0.99\times$ & \textbf{80.71} \\
                     & MSc-VBSWC & $1.04\times$ & $\mathbf{0.70}\times$ & $\mathbf{0.66}\times$ & $\mathbf{0.88}\times$ & 80.56 \\
        \midrule
        RegNetY-16GF \cite{regnet} & SSc-FBS   & $\mathbf{1.00}\times$ & $1.00\times$    & $1.00\times$    & $1.00\times$    & 79.64 \\
                     & MSc-FBS   & $2.29\times$ & $1.15\times$ & $1.00\times$    & $1.11\times$ & 80.52 \\
                     & MSc-VBS   & $1.11\times$ & $0.77\times$ & $0.77\times$ & $0.79\times$ & \textbf{80.96} \\
                     & MSc-VBSWC & $1.45\times$ & $\mathbf{0.69}\times$ & $\mathbf{0.66}\times$ & $\mathbf{0.72}\times$ & 80.64 \\
        \midrule
        EfficientNet-B3  \cite{efficientnet}& SSc-FBS   & $\mathbf{1.00}\times$ & $1.00\times$    & $1.00\times$    & $1.00\times$    & 81.20 \\
                        & MSc-FBS   & OOM & -            & -            & -            & - \\
                        & MSc-VBS   & $1.01\times$ & $0.66\times$ & $0.75\times$ & $0.78\times$ & 81.47 \\
                        & MSc-VBSWC & $1.03\times$ & $\mathbf{0.58}\times$ & $\mathbf{0.65}\times$ & $\mathbf{0.64}\times$ & \textbf{81.86} \\
         \bottomrule[1.5pt]
    \end{tabular}
    }
\caption{\textbf{Training models with multi-scale samplers reduces compute and training time while improving accuracy.} We train multiple CNN architectures using single-scale and multi-scale samplers on the ImageNet dataset. Multi-scale samplers are able to consistently match the performance of the single-scale models while reducing training time, optimization updates, and FLOPs.}
\label{table:all_models}
\end{table*}

\section{Results on Additional Models and Tasks}
\label{sec:generic}
In Section \ref{sec:why_msc_samplers}, we analyzed the efficacy of multi-scale samplers using ResNet-101 on the ImageNet dataset. In this section, we validate that the benefits of multi-scale training extend to other architectures and tasks. 

\subsection{Different image classification models}
We study three different models on the ImageNet dataset using different samplers (Section \ref{sec:overview}): (1) ResNet-50, (2) Se-ResNet-50 \cite{squeeze_excitation}, (3) RegNetY-16GF \cite{regnet}, and (4) EfficientNet-B3 \cite{efficientnet}. Results are given in Table \ref{table:all_models}. We observe that multi-scale samplers consistently improve the performance of different models and accelerate training. This is in conjunction with our observations in Section \ref{ssec:faster_training}. We provide additional results for vision transformer \cite{vit} and Swin transformer \cite{swin} models in the Appendix Section \ref{appendix:vit}.

\begin{table*}[t!]
    \centering
    \resizebox{2\columnwidth}{!}{
    \begin{tabular}{llcccccc}
        \toprule[1.5pt]
        \textbf{Pre-training Sampler} & \textbf{Mask R-CNN Sampler} & \textbf{Peak GPU Memory} & \textbf{Training FLOPs} & \textbf{Optimization updates} & \textbf{Training time} & \textbf{bbox mAP} & \textbf{segm mAP}\\
        \midrule[1.25pt]
        \multirow{3}{*}{SSc-FBS}   & SSc-FBS     & $\mathbf{1.00}\times$ & $1.00\times$    & $1.00\times$    & $1.00\times$ & 42.91 & 38.10 \\
           & MSc-VBS     & $3.44\times$ & $0.70\times$     & $0.63\times$    & $0.72\times$   & 46.23 & 41.27 \\
           & MSc-VBSWC   & $3.35\times$ & $\mathbf{0.63}\times$    & $\mathbf{0.54}\times$    & $\mathbf{0.66}\times$ & \textbf{46.48} & \textbf{41.28} \\
        \midrule
        \multirow{3}{*}{MSc-VBS}   & SSc-FBS     & $\mathbf{1.09}\times$ & $1.00\times$    & $1.00\times$    & $1.00\times$ & 43.19 & 38.30 \\
           & MSc-VBS     & $3.29\times$ & $0.70\times$     & $0.63\times$    & $0.72\times$ & \textbf{46.82} & \textbf{41.67} \\
           & MSc-VBSWC   & $2.68\times$ & $\mathbf{0.63}\times$    & $\mathbf{0.54}\times$    & $\mathbf{0.66}\times$ & 46.02 & 41.12 \\
        \midrule
        \multirow{3}{*}{MSc-VBSWC} & SSc-FBS     & $\mathbf{1.05}\times$ & $1.00\times$    & $1.00\times$    & $1.00\times$ & 42.71 & 38.12 \\
         & MSc-VBS     & $3.50\times$ & $0.7\times$     & $0.63\times$    & $0.72\times$ & \textbf{45.46} & 40.60 \\
         & MSc-VBSWC   & $3.30\times$ & $\mathbf{0.63}\times$    & $\mathbf{0.54}\times$    & $\mathbf{0.66}\times$ & 45.42 & \textbf{40.74} \\
        \midrule
          \multirow{3}{*}{None}       & SSc-FBS     & $\mathbf{0.91}\times$ & $1.00\times$    & $1.00\times$    & $1.00\times$ & 37.91 & 34.21 \\
                 & MSc-VBS     & $3.46\times$ & $0.7\times$     & $0.63\times$    & $0.74\times$ & \textbf{40.83} & \textbf{36.74} \\
                 & MSc-VBSWC   & $3.38\times$ & $\mathbf{0.63}\times$    & $\mathbf{0.54}\times$    & $\mathbf{0.68}\times$ & 39.00 & 35.74 \\
         \bottomrule[1.5pt]
    \end{tabular}
    }
\caption{\textbf{Training with multi-scale samplers increases mAP while decreasing training time and compute.} We report bounding box and instance segmentation mAP@IoU of 0.50:0.05:0.95 for a Mask R-CNN \cite{mask_rcnn} model with a ResNet-101 backbone. Compared to single-scale training, multi-scale samplers achieve better performance while reducing optimization updates by 46\% and training FLOPs by 37\%. Models in the lower part of the table were trained without pre-training the backbone model. We also note that effective batch sizes in object detection tasks are much smaller than classification tasks (e.g., 4 vs. 256 per GPU). Consequently, effective batch sizes at high resolutions are similar to the batch sizes at the base/reference resolution, causing multi-scale variable batch size samplers to behave like multi-scale fixed batch size samplers. Hence, we observe a large peak GPU memory.}
\label{table:resnet101_mask_rcnn_map}
\end{table*}

\subsection{Object Detection with Mask R-CNN}
We study the effect of different samplers on a standard detection model, Mask R-CNN \cite{mask_rcnn} with a ResNet-101 image backbone. We train and evaluate the model on the MS-COCO dataset \cite{ms_coco}. To understand the effect of different samplers, we train the Mask R-CNN model with and without pre-training across different sampler configurations. 

The results are summarized in Table \ref{table:resnet101_mask_rcnn_map}. We first note that batch sizes for object detection tasks are typically much smaller than for classification tasks (e.g., a common batch size per GPU for object detection is 4, whereas 256 is commonly used for classification). Therefore, at high resolutions, the effective batch size for multi-scale samplers is similar to the batch size of the base/reference resolution, causing multi-scale variable batch samplers to behave like the multi-scale fixed batch size sampler. Therefore, we observe a larger peak GPU memory when training object detection/instance segmentation tasks with variable batch samplers as compared to classification tasks (Table \ref{table:all_models}).

Our experiments show that the Mask R-CNN model with a multi-scale sampler significantly improves performance both when training from scratch and when pre-training. For example, a pre-trained MSc-VBS model attains a 0.47 mAP in detection when training the Mask R-CNN model with MSc-VBS, compared to 0.43 when training the Mask R-CNN model with SSc-FBS. We also observe a commensurate improvement in instance segmentation mAP, where training with MSc-VBS yields an mAP of 0.42, while training with SSc-FBS yields a 0.38 mAP. In addition to substantially improving the mAP, training with multi-scale samplers significantly reduces training time and compute. We also note that training with multi-scale samplers improves performance for both detection and instance segmentation tasks regardless of the pre-training method. Our results are consistent with large-scale jittering data augmentation \cite{ghiasi2021simple} for object detection models. However, a key difference between models trained with large-scale jittering and multi-scale samplers is that large-scale jittering uses a fixed batch size during training. As a result, training is slower. On the other hand, multi-scale samplers serve the dual purpose of improving accuracy as well as training efficiency.

\section{Conclusion}

We present an empirical study of multi-scale samplers for training deep neural networks in visual recognition tasks. Compared to single-scale training,  we showed that multi-scale training does not lower accuracy. Moreover, we showed that multi-scale training reduces training time and FLOPs, enhances robustness, acts as a regularizer, and results in better-calibrated models. We analyzed three different sampling strategies, including a novel curriculum-based multi-scale sampler that enforces a gradual increase in the resolution of images during the course of training, while adapting the batch size for smarter utilization of compute and memory at different stages of learning. Based on our experiments and analysis, we have demonstrated that using multi-scale samplers yields significant advantages compared to single-scale samplers in various aspects of training and model properties. Finally, we extended our study to other network architectures and visual recognition tasks, namely object detection and instance segmentation, demonstrating similar benefits of multi-scale training in these settings.

 In this work, we have shown that training with multi-scale samplers can reduce training compute of moderately-sized networks without compromising performance. Moreover, for a given architecture and task, our experiments maintained training hyperparameters across the different samplers---including learning rate schedules and optimizers. Thus, we have shown that multi-scale samplers can serve as more robust and efficient drop-in replacements for single-scale samplers.

 Overall, our work provides a solid foundation for further exploration and development of multi-scale samplers in the context of deep neural network training. Extensions to this work should investigate large-scale models (e.g., CLIP \cite{clip}). Related to this, training with multi-modal multi-task data, wherein different modalities and batch sizes are consumed, is a timely and exciting future direction. We hope that this research will encourage others to build on our findings and elaborate on the limitations and future directions.

{\small
\bibliographystyle{ieee_fullname}
\bibliography{egbib}
}

\clearpage
\appendix
\section{Appendix}

\subsection{Additional Architectures}
\subsubsection{Vision Transformers}\label{appendix:vit}
Our analysis of multi-scale samplers has centered around CNNs. In this section, we show that multi-scale samplers improves training efficiency of vision transformers while being competitive to SSc-FBS. We train ViT-B \cite{vit} and Swin-S \cite{swin} models using single and multi-scale samplers. Results are summarized in Table \ref{table:vit_models}. We observe that training ViT-B with MSc-VBS reduces compute by 24\% while improving accuracy. MSc-VBS can also be used to train Swin-S more efficiently with comparable accuracy. We observe a larger drop in accuracy when training ViT-B and Swin-S with MSC-VBSWC. We hypothesize that the smaller input resolutions used by MSc-VBSWC at the start of training adversely affect the patch embeddings. We used a standard patch size of $16\times 16$ and a minimum resolution of $128\times 128$ with an initial compression factor of $\rho_0=0.75$. Hence, the smallest input resolution at the start of training was $96\times 96$ for MSc-VBSWC. While MSc-VBSWC may still be used to train a vision transformer, larger min/max spatial resolutions may be required and is the subject of future work. 

\begin{table*}[t!]
    \centering
    \resizebox{2\columnwidth}{!}{
    \begin{tabular}{llccccc}
        \toprule[1.5pt]
        \textbf{Model} & \textbf{Sampler} & \textbf{GPU Memory} &  \textbf{Training FLOPs} & \textbf{Optimization updates} & \textbf{Training time} & \textbf{Top-1 Accuracy}\\
        \midrule[1.25pt]
        ViT-B \cite{vit}      & SSc-FBS               & $\mathbf{1.00}\times$          & $1.00\times$       & $1.00\times$    & $1.00\times$    & 79.76 \\
                              & MSc-VBS               & $1.75\times$                   & $0.76\times$       & $0.76\times$    & $0.80\times$    & \textbf{80.12} \\
                              & MSc-VBSWC             & $2.19\times$                   & $\mathbf{0.67}\times$ & $\mathbf{0.67}\times$ & $\mathbf{0.74}\times$ & 78.44 \\
      \midrule
        Swin-S \cite{swin}    & SSc-FBS               & $\mathbf{1.00}\times$          & $1.00\times$       & $1.00\times$    & $1.00\times$    & \textbf{82.93} \\
                              & MSc-VBS               & $1.58\times$                   & $0.91\times$       & $0.76\times$    & $0.89\times$    & 82.35 \\
                              & MSc-VBSWC             & $1.56\times$                   & $\mathbf{0.82}\times$ & $\mathbf{0.67}\times$ & $\mathbf{0.81}\times$ & 81.78 \\
         \bottomrule[1.5pt]
    \end{tabular}
    }
\caption{\textbf{Training vision transformer models with multi-scale samplers reduces compute and training time without a significant drop in accuracy.} We train ViT-B \cite{vit} and Swin-S \cite{swin} using single-scale and multi-scale samplers on the ImageNet dataset. Training with multi-scale samplers reduces training time, optimization updates, and FLOPs. We observe a drop in accuracy when training with MSc-VBSWC which we hypothesize is due to the small training resolutions at the beginning of training which may affect the patch embeddings.}
\label{table:vit_models}
\end{table*}

\subsubsection{Lightweight Models}
Our analysis thus far has only considered relatively ``heavyweight'' networks. In Table \ref{table:lightweight_models}, we train MobileNetv1 \cite{Howard2017MobileNetsEC}, MobileNetv2 \cite{Sandler2018MobileNetV2IR}, and MobileNetv3-Large \cite{Howard2019SearchingFM} using single-scale and multi-scale samplers. We observe that training with multi-scale samplers can improve efficiency with similar or better accuracy.

\begin{table*}[t!]
    \centering
    \resizebox{2\columnwidth}{!}{
    \begin{tabular}{llccccc}
        \toprule[1.5pt]
        \textbf{Model} & \textbf{Sampler} & \textbf{GPU Memory} &  \textbf{Training FLOPs} & \textbf{Optimization updates} & \textbf{Training time} & \textbf{Top-1 Accuracy}\\
        \midrule[1.25pt]
        MobileNetv1 \cite{Howard2017MobileNetsEC}     & SSc-FBS               & $\mathbf{1.00}\times$          & $1.00\times$       & $1.00\times$    & $1.00\times$    & 73.92 \\
                              & MSc-VBS               & $1.01\times$                   & $0.77 \times$       & $0.77 \times$    & $0.98\times$    & \textbf{74.16} \\
                              & MSc-VBSWC             & $1.24\times$                   & $\mathbf{0.70}\times$ & $\mathbf{0.66}\times$ & $\mathbf{0.97}\times$ & 74.05 \\
      \midrule
        MobileNetv2 \cite{Sandler2018MobileNetV2IR}      & SSc-FBS               & $\mathbf{1.00}\times$          & $1.00\times$       & $1.00\times$    & $1.00\times$    & \textbf{73.36} \\
                              & MSc-VBS               & $1.21\times$                   & $0.77\times$       & $0.77\times$    & $0.98\times$    & 73.06 \\
                              & MSc-VBSWC             & $1.13\times$                   & $\mathbf{0.70}\times$ & $\mathbf{0.66}\times$ & $\mathbf{0.91}\times$ & 73.09 \\
      \midrule
        MobileNetv3 \cite{Howard2019SearchingFM}      & SSc-FBS               & $\mathbf{1.00}\times$          & $1.00\times$       & $1.00\times$    & $1.00\times$    & 74.74 \\
                              & MSc-VBS               & $1.19\times$                   & $0.77\times$       & $0.77\times$    & $0.94\times$    & \textbf{75.24} \\
                              & MSc-VBSWC             & $1.42\times$                   & $\mathbf{0.71}\times$ & $\mathbf{0.66}\times$ & $\mathbf{0.88}\times$ & 74.77 \\
         \bottomrule[1.5pt]
    \end{tabular}
    }
\caption{\textbf{Multi-scale samplers reduce training FLOPs of lightweight CNNs without a significant drop in performance.} We train MobileNet models on ImageNet using the recipes in \cite{cvnets}. We observe that training lightweight networks with multi-scale samplers produces models with accuracies that are competitive to the SSc-FBS model while being more efficient to train.}
\label{table:lightweight_models}
\end{table*}

\subsection{Additional Rresults}
\subsubsection{Number of GPUs and Uncertainty Measurements}\label{appendix:gpus}
The focal point of this paper has been on the performance and efficiency gains of multi-scale variable batch size samplers. These samplers typically assume multi-GPU training, however, they may still be utilized using a single GPU. Table \ref{table:resnet101_mean_std} compares the top-1 (\%) accuracy of a ResNet-101 model trained using single and multi-scale samplers on platforms with either 1 or 4 GPUs. For each sampler and GPU configuration, we report the mean and standard deviation for three seeds (due to computational constraints, we are unable to provide such measurements for all experiments). We observe that, for a given number of GPUs, the multi-scale sampler performance is slightly lower than the single-scale sampler performance when trained on a single GPU. However, the use of multiple GPUs for training enables multi-scale samplers to exceed the performance of the single-scale sampler. We hypothesize that the aggregation of gradients across different batch sizes and resolutions facilitates learning.

\begin{table}[t!]
    \small
    \centering
    \resizebox{\columnwidth}{!}{
    \begin{tabular}{ccc}
        \toprule[1.5pt]
        \# GPUs & Sampler & Top-1 Accuracy (\%) \\
        \hline
        4 & SSc-FBS & $81.03 \pm 0.32$ \\
        & MSc-VBS & $81.79 \pm 0.09$ \\
        & MSc-VBSWC & $81.56 \pm 0.10$ \\
        \hline
        1 & SSc-FBS & $80.15 \pm 0.11$ \\
        & MSc-VBS & $79.01 \pm 0.23$ \\
        & MSc-VBSWC & $79.68 \pm 0.50$ \\
        \bottomrule
    \end{tabular}
    }
\caption{ResNet-101 ImageNet-1k top-1 accuracy aggregated over three seeds. 4-GPU models were trained for 600 epochs while 1-GPU models were trained for 150 epochs.}
\label{table:resnet101_mean_std}
\end{table}

\subsection{Entropy Details}

In Section, \ref{sec:robustness} we measured entropy skewness as a means of quantifying the rate at which the model becomes more confident in its predictions. In this section, we expand on this metric and provide additional background. For a given image, $I$, a classification network's softmax output, $\gN(I) = [\hat{p}_0, \ldots, \hat{p}_{K-1}]$, is often interpreted as a categorical distribution over the possible image categories. The uncertainty in this prediction can be quantified by computing the entropy of this distribution, given by Equation (\ref{eq:sm_logit_entropy}).

\begin{align}\label{eq:sm_logit_entropy}
    \gH(\gN(I)) = -\sum_{i=0}^{K-1} \hat{p}_i \log \hat{p}_i.
\end{align}

\begin{figure}[t!]
    \centering
    \begin{subfigure}[b]{0.45\columnwidth}
        \centering
        \includegraphics[width=\columnwidth]{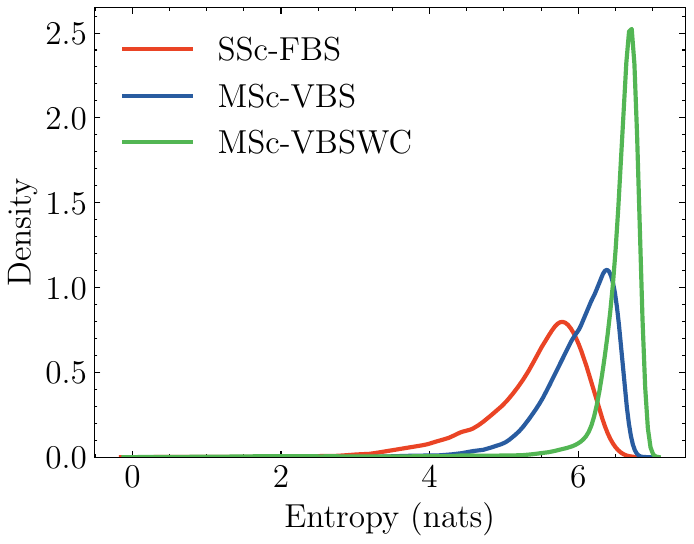}
        \caption{Epoch 0}
        \label{fig:resnet101_entropy_dist_0}
    \end{subfigure}
    \begin{subfigure}[b]{0.45\columnwidth}
        \centering
        \includegraphics[width=\columnwidth]{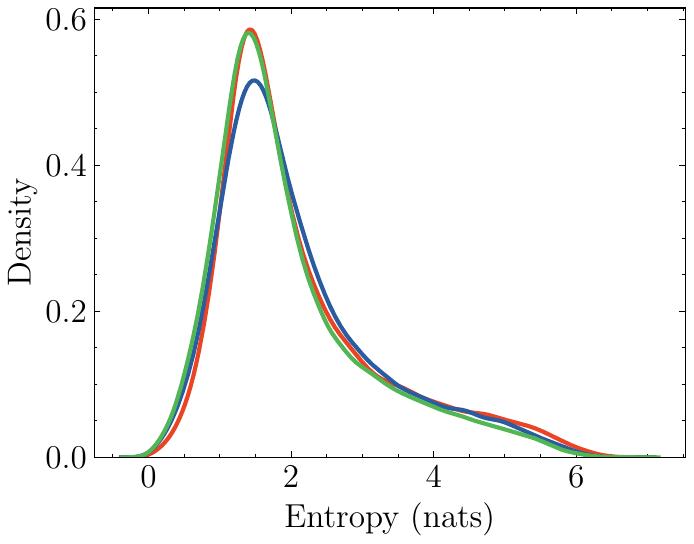}
        \caption{Epoch 599}
        \label{fig:resnet101_entropy_dist_599}
    \end{subfigure}
    \caption{\textbf{Entropy decreases throughout training.} At the start of training, the (softmax) distributions output by a classification network will have a high entropy (a). Over the course of training, the entropy gradually decreases (b). Entropy densities are constructed at the end of the first epoch (a) and at the end of the final epoch (b). At the end of training, single-scale and multi-scale samplers produce models with similar confidence levels.}
    \label{fig:resnet101_entropy_dist}
\end{figure}

Computing the entropy for every image in the validation set allows us to construct an empirical distribution over the possible entropy values. As depicted in Figure \ref{fig:resnet101_entropy_dist_0}, the entropy at the beginning of training is skewed towards the left (higher mass at large entropy values). At the end of training, the distribution is skewed towards the right (higher mass at lower entropy values), as depicted in Figure \ref{fig:resnet101_entropy_dist_599}. We therefore propose to investigate the rate at which this skewness shift occurs. More formally, let $I_j, j = 1, \ldots, N_{val}$ be the  images in the validation set and $\gE = \{\gH(\gN(I_j)) : j = 1, \ldots, N_{val}\}$ be the set of entropies for all images in the validation set. Letting $\mu$ and $\sigma$ be the sample mean and standard deviation of $\gE$, the skewness of the empirical distribution induced by $\gE$ is given by Equation (\ref{eq:skewness}).

\begin{align}\label{eq:skewness}
    \textrm{Skewness}(\gE) &= \frac{1}{N_{val}}\sum_{j=1}^{N_{val}} \left(
    \frac{\gH(\gN(I_j)) - \mu}{\sigma}
    \right)^3
\end{align}

A distribution that is skewed left will have a negative skewness--similar to the distribution in Figure \ref{fig:resnet101_entropy_dist_0} at the start of training--and a distribution that is skewed right will have a positive skewness. At the end of each training epoch, we compute the skewness of the model's entropy distribution. By monitoring the shift from negative to positive skewness values of these distributions, we can gain some insight into the rate at which models become more confident in their predictions.

\subsection{MSc-VBSWC Details}\label{appendix:vbswc_details}
In this section, we describe how we extended the MSc-VBS sampler with a curriculum to obtain the MSc-VBSWC sampler. As discussed in Section \ref{sec:overview}, at each epoch, MSc-VBSWC considers a dynamic set of possible spatial resolutions, $s(e) = \{(h_1(e), w_1(e)), \ldots, (h_n(e), w_n(e))\}$.  At each training iteration of epoch $e$, a spatial resolution is sampled from $s(e)$. Contrary to the MSc-VBS sampler that samples from a fixed set of spatial resolutions, $\gS$, in Msc-VBSWC we begin training with a ``compressed'' version of $\gS$ in which the spatial resolutions have been scaled by a factor. At epoch $e$, the set of resolutions to sample from is $s(e) = \{(\rho(e) \cdot H_1, \rho(e)\cdot W_1), \ldots, (\rho(e) \cdot H_n, \rho(e)\cdot W_n)\}$ where $\rho(e)$ is a monotonically increasing function such that $\rho(0) = \rho_0 \in (0,1]$,  $\tau \in (0,1]$, and $\rho(\tau E) = 1$. $\rho_0$ represents the initial compression factor and $\tau$ determines the fraction of epochs it takes for $s(e)$ to expand to $\gS$. For example, for $\rho_0=0.75$, $\tau=0.5$, and $E=600$, training begins by sampling from $s(0) = \{(0.75 H, 0.75W) : (H,W) \in \gS\}$, and for $e \geq 300$, $s(e) \equiv \gS$. For all MSc-VBSWC experiments, we use $\rho_0 = 0.75$ and $\tau = 0.5$.

\begin{figure}[t]
\begin{center}
   \includegraphics[width=0.9\linewidth]{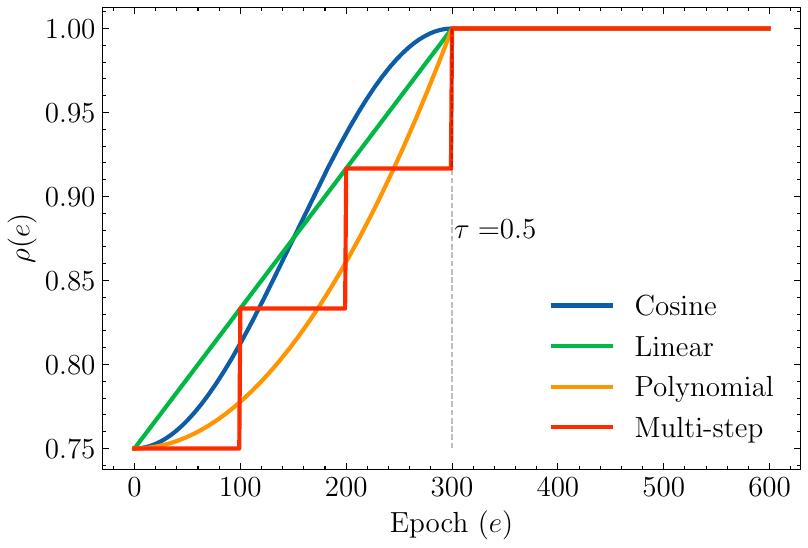}
\end{center}
   \caption{\textbf{MSc-VBSWC expansion schedules.} We experiment with four expansion schedules, $\rho(e)$, for MSc-VBSWC that control the rate at which spatial resolutions expand throughout training. We use initial compression factor $\rho_0=0.75$, and expansion period $\tau=0.5$.}
\label{fig:vbswc_rho_vs_e}
\end{figure}

We consider four schedules for $\rho(e)$: 1) linear, 2) cosine, 3) polynomial, and 4) multi-step. A plot of these schedules is provided in Figure \ref{fig:vbswc_rho_vs_e}. Next, we train a ResNet-101 architecture on ImageNet using each of the four samplers and summarize the results in Table \ref{tab:resnet101_vbswc_schedules}. Due to the stronger accuracy-efficiency tradeoff of the cosine schedule, we adopt the cosine schedule for all of our MSc-VBSWC experiments.

\begin{table*}[t!]
    \small
    \centering
    \begin{tabular}{lccccc}
        \toprule[1.5pt]
        \textbf{Sampler} & \textbf{Peak GPU Memory} &  \textbf{Training FLOPs} & \textbf{Optimization updates} & \textbf{Training time} & \textbf{Top-1 Accuracy}\\
        \midrule[1.25pt]
        SSc-FBS   & $\mathbf{1.00\times}$    & $1.00\times$    & $1.00\times$    & $1.00\times$    & 81.31 \\
        \midrule
        MSc-VBSWC (Linear) & $1.72\times$ & $0.70\times$ & $0.66\times$ & $0.91\times$ & 81.34 \\
        MSc-VBSWC (Cosine) & $1.75\times$ & $0.70\times$ & $0.66\times$ & $\mathbf{0.84\times}$ & \textbf{81.53} \\
        MSc-VBSWC (Polynomial) & $1.61\times$ & $\mathbf{0.67\times}$ & $\mathbf{0.63\times}$ & $0.88\times$ & 81.29 \\
        MSc-VBSWC (Multi-step) & $1.32\times$ & $\mathbf{0.67\times}$ & $0.64\times$ & $0.87\times$ & 81.22 \\
         \bottomrule[1.5pt]
    \end{tabular}
    \caption{\textbf{MSc-VBSWC with a cosine schedule has a stronger accuracy-efficiency trade-off}. We train a ResNet-101 architecture on ImageNet with linear, cosine, polynomial, and multi-step expansion schedules. Due to its stronger accuracy-efficiency trade-off, we adopt the cosine schedule for all of our MSc-VBSWC experiments.}
    \label{tab:resnet101_vbswc_schedules}
\end{table*}

\subsection{Training Details}
In this section we provide additional training details of our experiments. For a particular model, all sampling schedules (SSc-FBS, MSc-FBS, MSc-VBS, MSc-VBSWC) are trained using the same recipe. Hence, multi-scale samplers are drop-in replacements for single-scale samplers, requiring no modifications to the existing SSc-FBS recipe. All models are trained with the corresponding recipes implemented in the CVNets library \cite{cvnets}. RegNet \cite{regnet} models are trained with the ResNet recipe. Though CVNets recipes consider the exponential moving average (EMA) of training checkpoints, all of our experimental results are reported for the non-EMA model. All of our CNN architectures are trained on a single node with four NVIDIA A100 GPUs. Vision transformer models are trained on a single node with eight NVIDIA A100 GPUs.

When training with a multi-scale sampler (MSc-FBS, MSc-VBS, MSc-VBSWC), we must first specify a ``reference'' batch shape as discussed in Section \ref{sec:overview}. Additionally, we define a minimum spatial resolution, $(H_1, W_1)$, and a maximum, $(H_n, W_n)$. To construct the spatial resolutions in $\gS$, we interpolate between the minimum and maximum spatial resolutions, imposing divisibility by either 8 or 32. The batch sizes in $\gS$ are computed based on the batch size of the reference batch shape. We report the reference batch shapes as well as the min/max spatial resolution used for each of our models and samplers in Table \ref{table:msc_reference_batch_shapes}.

\begin{table}[t!]
    \centering
    \resizebox{\columnwidth}{!}{
    \begin{tabular}{lcccc}
        \toprule[1.5pt]
        \textbf{Model} & \textbf{MSc-VBS Ref.} & \textbf{MSc-VBSWC Ref.} & \textbf{Min (H,W)} & \textbf{Max (H,W)}\\
        \midrule[1.25pt]
        ResNet \cite{resnet} & (256, 3, 224, 224) & (512, 3, 160, 160) & 128 & 320 \\
        RegNet \cite{regnet} & (256, 3, 224, 224) & (512, 3, 160, 160) & 128 & 320 \\
        EfficientNet \cite{regnet} & (256, 3, 300, 300) & (256, 3, 300, 300) & 160 & 448 \\
        ViT \cite{vit} & (256, 3, 224, 224) & (256, 3, 224, 224) & 128 & 320 \\
        Swin \cite{swin} & (256, 3, 224, 224) & (256, 3, 224, 224) & 128 & 320 \\
        Mask R-CNN \cite{mask_rcnn} & (4, 3, 1024, 1024) & (4, 3, 1024, 1024) & 512 & 1280 \\
        \bottomrule
    \end{tabular}
    }
    \caption{\textbf{Reference batch shapes and min/max spatial resolutions for MSc-VBS and MSc-VBSWC}. The reference batch shapes for each sampler are used to determine the batch size of the spatial resolutions sampled at each training iteration. Reference batch shapes maintain a similar compute within each model and are reported as $(B, C, H, W)$. Spatial resolutions at each iteration are sampled from resolutions interpolated between the min/max spatial resolutions.}
    \label{table:msc_reference_batch_shapes}
\end{table}

\end{document}